\documentclass[lettersize,journal]{IEEEtran}
\usepackage{amsmath,amsfonts}
\usepackage{algorithmic}
\usepackage{array}
\usepackage[caption=false,font=normalsize,labelfont=sf,textfont=sf]{subfig}
\usepackage{textcomp}
\usepackage{stfloats}
\usepackage{url}
\usepackage{verbatim}
\usepackage{graphicx}
\usepackage{subfiles}
\usepackage{amsmath,amssymb}
\usepackage{xpatch} 
\usepackage[utf8]{inputenc} 
\usepackage[T1]{fontenc}
\usepackage{cite}
\usepackage{booktabs}       
\usepackage{amsfonts}       
\usepackage{nicefrac}       
\usepackage{microtype}      %
\usepackage{adjustbox}
\usepackage{amsmath}
\usepackage{sidecap}
\usepackage{multirow}
\usepackage{verbatim}
\usepackage[table]{xcolor}

\newcommand{\mypar}[1]{\vspace{.25em}\noindent\textbf{#1}~}

\newcommand{\mr}[1]{\mathrm{#1}}

\newcommand{\ppm}{\,\scriptsize$\pm$}
\usepackage[pagebackref,breaklinks,colorlinks]{hyperref}

\newcommand{\XX}{\mathcal{X}}
\newcommand{\YY}{\mathcal{Y}}
\newcommand{\TT}{\mathcal{T}}

\newcommand{\Loss}{\mathcal{L}}

\def\BibTeX{{\rm B\kern-.05em{\sc i\kern-.025em b}\kern-.08em
    T\kern-.1667em\lower.7ex\hbox{E}\kern-.125emX}}
\usepackage{balance}

\begin{document}
\title{TFS-ViT: Token-Level Feature Stylization for Domain Generalization}

\author{\IEEEauthorblockN{\textbf{Mehrdad Noori}\IEEEauthorrefmark{1}\IEEEauthorrefmark{2},
\textbf{Milad Cheraghalikhani}\IEEEauthorrefmark{1},
\textbf{Ali Bahri}, 
\textbf{Gustavo A. Vargas Hakim}, \\
\textbf{David Osowiechi}, 
\textbf{Ismail Ben Ayed}, 
\textbf{Christian Desrosiers}} 
\\ \vspace{0.3cm}\IEEEauthorblockA{LIVIA, ÉTS Montreal, Quebec, Canada}
\\ \IEEEauthorblockA{International Laboratory on Learning Systems (ILLS)}
\thanks{\IEEEauthorrefmark{1}Equal contribution}
\thanks{\IEEEauthorrefmark{2}Corresponding author: M. Noori (email: mehrdad.noori.1@ens.etsmtl.ca)}
}

\markboth{}{}

\maketitle

\begin{abstract}
Standard deep learning models such as convolutional neural networks (CNNs) lack the ability of generalizing to domains which have not been seen during training. This problem is mainly due to the common but often wrong assumption of such models that the source and target data come from the same i.i.d. distribution. Recently, Vision Transformers (ViTs) have shown outstanding performance for a broad range of computer vision tasks. However, very few studies have investigated their ability to generalize to new domains. This paper presents a first Token-level Feature Stylization (TFS-ViT) approach for domain generalization, which improves the performance of ViTs to unseen data by synthesizing new domains. Our approach transforms token features by mixing the normalization statistics of images from different domains. We further improve this approach with a novel strategy for attention-aware stylization, which uses the attention maps of class (CLS) tokens to compute and mix normalization statistics of tokens corresponding to different image regions. The proposed method is flexible to the choice of backbone model and can be easily applied to any ViT-based architecture with a negligible increase in computational complexity. Comprehensive experiments show that our approach is able to achieve state-of-the-art performance on five challenging benchmarks for domain generalization, and demonstrate its ability to deal with different types of domain shifts. The implementation is available at \href{https://github.com/Mehrdad-Noori/TFS-ViT\_Token-level\_Feature\_Stylization}{this repository}.
\end{abstract}

\begin{IEEEkeywords}
Deep Learning, Domain Generalization, Feature Stylization, Vision Transformer.
\end{IEEEkeywords}

\section{Introduction}
\label{sec:intro}
Deep learning models like convolutional neural networks (CNNs), and more recently Vision Transformers (ViT), have enabled unprecedented progress in computer vision, achieving state-of-art performance on various tasks such as classification, semantic segmentation and object detection. However, most of these models rely on the naive assumption that the data used for training (the source domain) and the one encountered after deployment (the target domain) come from the same distribution. As they are not designed to tackle distribution shifts, the performance of such models typically degrades when out-of-distribution (OOD) data is encountered~\cite{recht2019imagenet,hendrycks2019benchmarking}. Domain adaptation (DA) approaches~\cite{lu2020stochastic,saito2018maximum} attempt to solve this problem by adapting a model trained on source domain data to a known target domain. A major limitation of such approaches is that they need target data for adaptation, which is not always available in practice. Moreover, adapting the source-trained model to each new target domain also incurs additional costs in terms of computations.  

Domain generalization (DG)~\cite{blanchard2011generalizing} seeks to overcome the domain shift problem in a different way: training a model with data from one or multiple source domains so that it can generalize to OOD data from any target domain. In recent years, a plethora of methods have been proposed for this challenging problem~\cite{zhou2022domain,wang2022generalizing}, exploiting various strategies including domain alignment~\cite{hu2020domain,mahajan2021domain,li2020domain}, meta-learning~\cite{li2018learning,balaji2018metareg}, data augmentation~\cite{shi2020towards,shankar2018generalizing}, ensemble learning~\cite{zhou2021domain}, self-supervised learning~\cite{carlucci2019domain,albuquerque2020improving} and regularization~\cite{huang2020self, cha2021swad}. While many of these methods have shown promising results using CNN architectures, very few have investigated the potential of ViTs for DG~\cite{sultana2022self}. One key reason for this is the more limited understanding of how ViTs learn compared to CNNs. For instance, it is well known that the first layers of CNN architectures like ResNet encode domain-specific features, while those closer to the output capture features that are more related to class~\cite{zhou2021domain}. This knowledge enables the development of efficient DG strategies, for example, augmenting the features in early network layers while enforcing the classification output to be consistent. 

Compared to CNNs, which mainly learn by recognizing and composing local patterns, ViTs can model global relationships using so-called multiheaded self-attention (MSA) layers~\cite{dosovitskiy2020image}. Although ViT features are harder to interpret than those learned by CNNs, attention maps in MSA layers offer a powerful way to analyze the relationships between different parts of an image and their link to semantic classes. In particular, attention maps to the class (CLS) token measure of the contribution of each region to predicting the class of an image. In a recent work~\cite{choi2022tokenmixup}, authors exploit the attention maps of ViTs in a token-level data augmentation method for classification. While it improved performance by maximizing the saliency of augmented tokens, this method was designed for standard supervised learning, and not for a DG setting where the model can encounter OOD data.

In this paper, we propose a novel domain generalization approach, called Token-level Feature Stylization (TFS-ViT), which improves the generalization performance of ViTs to OOD data by synthesizing new domains. The core idea of our approach is to augment token-level features by mixing the normalization statistics of images from different domains. This encourages the model to learn meaningful relationships between different parts of an image, which do not depend on the image's style. We improve this approach with an attention-aware stylization strategy that leverages the attention maps of class (CLS) tokens to compute and mix normalization statistics of tokens corresponding to different image regions. The proposed method is flexible to the choice of backbone model and can be easily applied to any ViT-based architecture with a negligible increase in computational complexity. 

Our contributions can be summarized as follows:
\begin{itemize}\setlength\itemsep{.3em}
\item We present a first token-level feature stylization approach for domain generalization in ViTs;
\item We extend this approach with a novel attention-aware stylization strategy that uses attention maps in MSA layers to guide the augmentation toward more important regions of the image;
\item We conduct extensive experiments on five challenging datasets, using different ViT architectures, and show our method to achieve state-of-art performance in most cases.  
\end{itemize}
The rest of the paper is organized in the following way. In the next section, we provide an overview of related works on DG and recent methods using ViTs for this task. Section~\ref{sec:method} then defines the DG problem addressed in this work, and presents our TFS-ViT approach for this problem. Section~\ref{sec:exp} describes the datasets and implementation details related to experiments. In Section~\ref{sec:res}, we present results evaluating the different components of our method and showing its advantage over existing approaches. Finally, we discuss the main results and possible future directions in Section~\ref{sec:conclusion}.

\section{Related Works}

\mypar{Domain Generalization} The problem of generalizing to OOD data was initially introduced by Blanchard et al. in 2011~\cite{blanchard2011generalizing} and has since then generated a growing interest in computer vision. The broad range of methods developed for this problem can mostly be grouped in seven categories based on domain alignment, meta-learning, data augmentation, ensemble learning, self-supervised learning, disentangled representation learning, and regularization. Domain alignment methods seek to learn domain-invariant representations by minimizing the difference among available source domains. This can be achieved in several ways, for instance by matching moments~\cite{peng2019moment}, using discriminant analysis~\cite{hu2020domain}, minimizing the maximum mean discrepancy (MMD)~\cite{liDomainGeneralizationAdversarial2018} or a contrastive loss~\cite{motiian2017unified}, as well as with domain-adversarial learning~\cite{li2018deep}. Meta-learning approaches for DG~\cite{li2018learning,balaji2018metareg} typically consider a bi-level optimization problem where a model is updated using meta-source domains so that the test error on a given meta-target domain is minimized. This meta-learning is often done using episodic training, and can update all parameters of a network~\cite{li2018learning} or a reduced set of regularization parameters~\cite{balaji2018metareg}. 

Data augmentation is another popular approach for DG, which simulates domain shifts during training in hope of making the model more robust to such shifts. This can be achieved using various strategies, including learnable augmentation, off-the-shelf style transfer and feature-based augmentation. The first strategy employs an augmentation network to generate images from training samples so that their joint distribution is different from those of existing source domains~\cite{carlucciHallucinatingAgnosticImages2019,zhouLearningGenerateNovel2020a}. On the other hand, data augmentation methods based on off-the-shelf style transfer try to map input images from one domain to another using techniques like Fourier-based augmentation~\cite{xu2023fourier} or to change the style of these images~\cite{yueDomainRandomizationPyramid2019}. This can be done, for example, using Adaptive Instance Normalization (AdaIN)~\cite{huangArbitraryStyleTransfer2017,somavarapuFrustratinglySimpleDomain2020}. While most data augmentation methods operate on pixels, feature-level augmentation techniques have also been proposed for DG~\cite{manciniRecognizingUnseenCategories2020,zhouMixStyleNeuralNetworks2021}. Such techniques are motivated by the observation that style-related information is often captured in statistics of CNN features~\cite{zhouMixStyleNeuralNetworks2021}. 

Ensemble learning approaches for DG try to increase the robustness to OOD data by training multiple domain-specific models~\cite{zhou2021domain}. The ensemble prediction for target domain examples can then be obtained as a weighted average of the individual models' predictions, with the weights measuring the similarity of the target sample to each source domain or the models' confidence~\cite{mancini2018best}. In contrast, self-supervised learning methods aim to learn representations that better generalize across domains by pre-training a model on some unsupervised auxiliary (pretext) tasks~\cite{carlucci2019domain,Kim_2021_ICCV}. Pretext tasks can be solving a jigsaw puzzle~\cite{carlucci2019domain,wang2020learning}, reconstructing an image with an autoencoder~\cite{ghifary2015domain}, or using a contrastive objective~\cite{Kim_2021_ICCV}. In a more recent study~\cite{zhang2023deep}, the connection between the information bottleneck principle and DG has been studied, leading to enhanced domain generalization through domain-invariant representation learning.     

Instead of directly learning a domain-invariant representation, disentangled representation learning approaches try to separate features in two groups encoding domain-specific and domain-invariant information~\cite{li2017deeper,chattopadhyay2020learning}. This can be achieved by learning domain-specific masks that can dynamically select relevant features for a given image of the target domain~\cite{chattopadhyay2020learning}. The last category of methods for DG regularize the training of a model to learn features which can better generalize across domains~\cite{cha2021swad,sagawa2019distributionally,sultana2022self}. Such methods typically extend the empirical risk minimization (ERM) approach~\cite{Gulrajani2021InSO} by adding a regularization objective, for example, based on distillation~\cite{sultana2022self}, dense stochastic weight averaging~\cite{cha2021swad} or distributionally robust optimization (DRO)~\cite{sagawa2019distributionally}.

\mypar{Vision transformers (ViTs)} The methods mentioned above are mostly based on CNN architectures. Despite the outstanding performance of ViTs for classification~\cite{dosovitskiy2020image}, object detection~\cite{Dai_2021_ICCV} and semantic segmentation~\cite{strudel2021segmenter}, very few works have explored their potential for domain generalization. Zhang et al. analyzed the robustness of ViTs to distribution shifts, and proposed a novel architecture based on self-supervised learning and information theory that better generalizes to data from unseen domains~\cite{zhang2021delving}. Recently, Sultana et al. proposed a Self-Distilled Vision Transformer (SDViT) for DG which employs auxiliary losses in intermediate transformer blocks to alleviate the problem of overfitting source domains~\cite{sultana2022self}. Our proposed method follows a different approach: designing an token-level features stylization strategy that exploits the information of attentions maps to effectively and efficiently synthesize new domains during training.

\section{Method}
\label{sec:method}

In this section, we first define the problem of domain generalization. We then detail our Token-Level Feature Stylization (TFS-ViT) method for DG and explain how attention maps in MSA layers can be used to further improve its performance.

\subsection{Problem Definition}

Referring to the input space as $\XX$ and the target space as $\YY$, we define a domain for a classification task as the joint distribution of $P_{\XX\YY}$ on $\XX\!\times\!\YY$. For a particular domain, the marginal distribution on $\XX$ is denoted as $P_{\XX}$, the posterior distribution of $\YY$ given $\XX$ as $P_{\YY|\XX}$, and the class-conditional distribution of $\XX$ given $\YY$ as $P_{\XX|\YY}$. In the standard DG setup, we have access to $M$ source domains that are related to one another but are not the same, $\mathcal{S} = \{S_i\}^{M}_{i=1}$. In other words, we proceed on the assumption that the joint distribution of each domain, $P^{(i)}_{\XX\YY}$, is unique in comparison to that of other domains, $P^{(i)}_{\XX\YY} \neq P^{(i')}_{\XX\YY}$ when $i \neq i'$. Each source domain consists of $N_i$ samples, $S_i = \{(x_j^{(i)}, y_j^{(i)} )\}_{j=1}^{N_i}$. $T = \{x_j^{\TT} \}_{j=1}^{N_\TT}$ represents the target domain, which has a joint distribution distinct from the one of the source domain. The goal is to predict the labels for target domain examples without having access to such examples. We thus try to minimize a loss function, $ \Loss : \YY\!\times\!\YY \to [0,\infty] $, to find the learning function $f : \XX \to \YY$ that best estimates $P_{\YY|\XX}$.

\begin{figure*}
  \centering
   \includegraphics[width=0.965\linewidth]{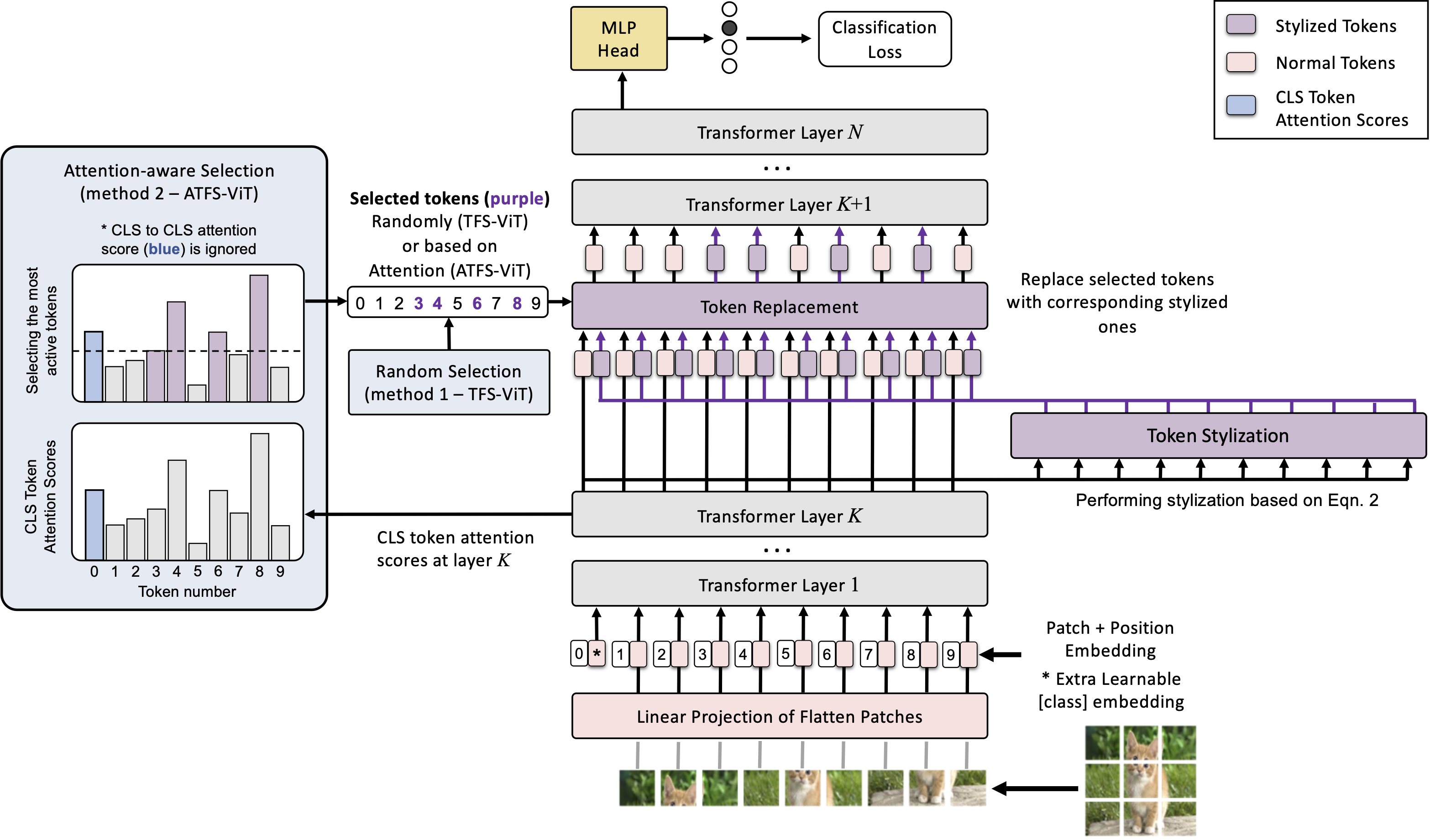}
  \caption{Overview of the proposed architecture for Token-level Feature Stylization (TFS-ViT).}
  \label{fig:tfs-arch}
\end{figure*}

\subsection{Token-Level Feature Stylization (TFS)}

Normalization-based feature stylization techniques, such as Adaptive Instance Normalization (AdaIN) \cite{huang2017arbitrary} and MixStyle~\cite{zhouMixStyleNeuralNetworks2021} have been shown to improve the generalization performance of CNNs. However, their potential  and effectiveness in the context of ViT models has not been explored. Motivated by the success of these techniques in CNNs and also by leveraging the sequential nature of ViTs, we propose a token-level feature stylization method, TFS-ViT, that is able to enhance the generalization capacity of ViTs on unseen domains. Our proposed method is designed to selectively stylize a subset of tokens at each layer, resulting in generating more divers samples during training. Figure~\ref{fig:tfs-arch} illustrates the overall architecture of TFS-ViT.

In the proposed method, we first estimate token-level statistics of the feature embedding in layer $k$, denoted by $x^k$, by computing the mean and standard deviation across token sequences: 
\begin{equation}
    \begin{split}
        \mu_{c}(x^k) & \, = \, \frac{1}{S} \sum_{s=1}^{S} x_{c,s}^k \\
        \sigma_{c}(x^k) & \, = \, \sqrt{ \frac{1}{S} \sum_{s=1}^{S} \big(x_{c,s}^k - \mu_{c}(x^k) \big)^2  }
    \end{split}
\end{equation}
where $S$ is the length of the token embedding sequence. We then randomly choose another sample $\Tilde{x}$ from the batch and synthesize a stylized version of $x^k$, denoted as $\phi(x^k)$, in the following manner:
\begin{equation}
    \begin{split}
        \gamma_{\mr{mix}} & \, = \, \alpha \sigma(x^k) \, + \, (1-\alpha) \sigma(\Tilde{x}^k)
        \\
        \beta_{\mr{mix}} & \, = \, \alpha \mu(x^k) \, + \, (1-\alpha) \mu(\Tilde{x}^k)
        \\
        \phi(x^k) & \, = \, \gamma_{\mr{mix}} \frac{x^k - \mu(x^k)}{\sigma(x^k)} \, + \, \beta_{\mr{mix}}
    \end{split}
\end{equation}
where mixing coeffcient $\alpha$ is sampled from the Beta distribution, $\alpha\!\sim\!\mr{Beta}(0.1, 0.1)$. Afterwards, to generate the input for the subsequent layer, we randomly choose a given number of tokens and replace their original feature, $x^k$, with their corresponding stylized version from $\phi(x^k)$. The percentage of replaced tokens is controlled by a hyper-parameter $d$. While training the network, at each iteration, we randomly choose $n$ layers from the total of $N$ layers that form the backbone and perform token-level stylization on those layers as described above. 

\begin{figure*}[t]
  \centering
   \includegraphics[width=.8\linewidth]{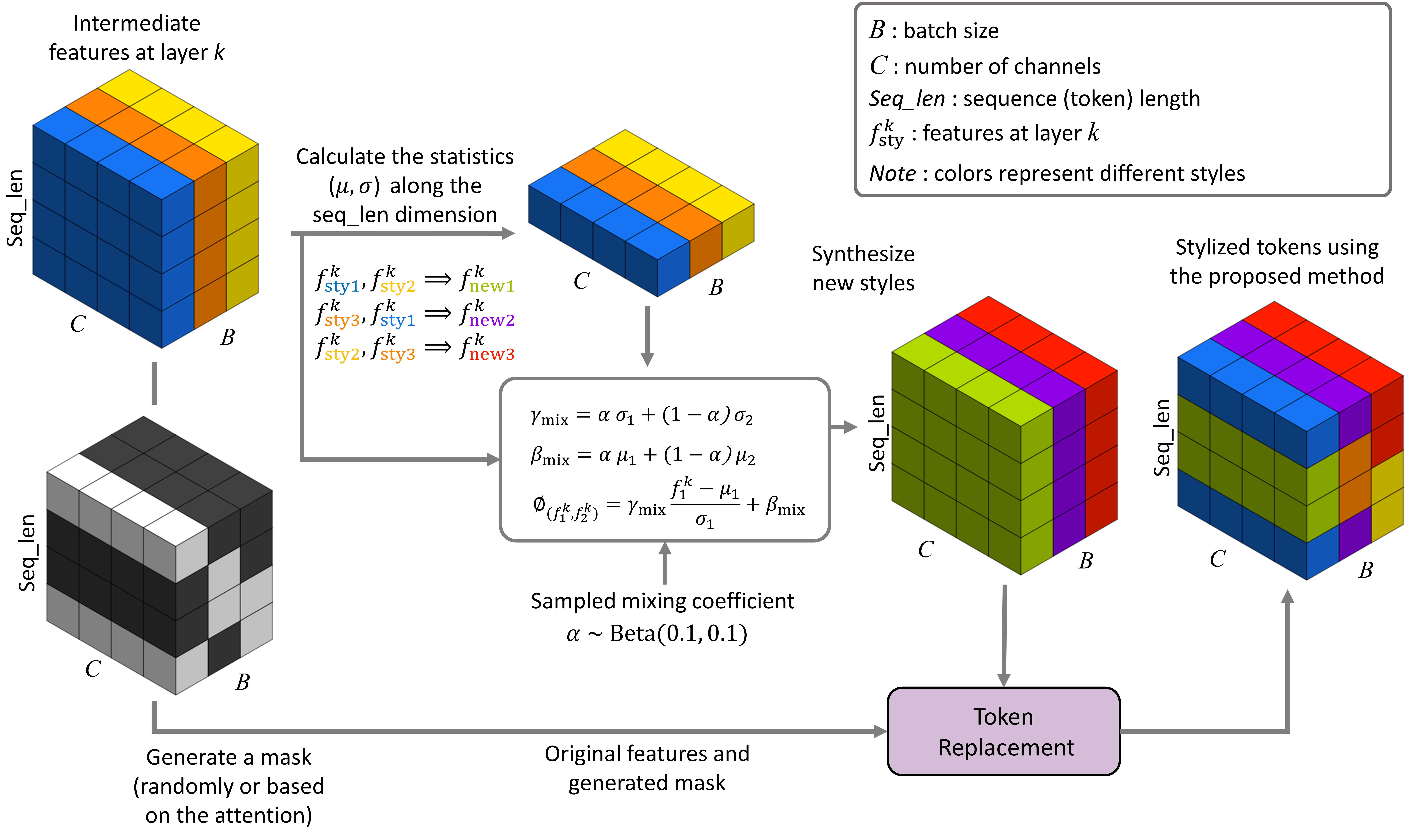}
   \caption{Synthesized features using our proposed method. Different colors denote different styles. By randomly selecting a subset of tokens to stylize at each layer, our method generates diverse samples while preserving the underlying structure of the tokens. This leads to forcing the network to only focus on the structure-related information which eventually results in improving the generalization performance. It is worth mentioning that we perform our stylization method on multiple layers of the ViT network.}
   \label{fig:stylization}
\end{figure*}

The detailed process of our TFS-ViT method is illustrated in Figure~\ref{fig:stylization}. The reason for stylizing some of the tokens while leaving others unchanged is to increase the diversity of the generated samples during training. By randomly selecting a subset of tokens to stylize at each layer, we are effectively creating new combinations of stylized tokens while also maintaining the underlying structure of the input tokens\footnote{We simulate our proposed stylization method by using an Encoder-Decoder ViT and visualize the effect of this selection and replacement of tokens in Figure 1 and Figure 2 of the supplementary materials.}. This cannot be achieved so easily in a CNN architecture. Our approach, specifically designed for ViTs, allows exploring a wider range of feature distributions, thereby increasing the model's capacity to generalize to unseen domains. Our method not only proves to be effective in DG settings (Section~\ref{subsection:Comparison}), but also enhances in-domain performance, as evidenced by the results presented in Section~\ref{subsection:Further_Analysis}.

\subsection{Attention-Aware TFS}
One of the key aspects of ViTs that distinguishes them from traditional CNNs is their use of  self-attention. In ViTs, self-attention maps are employed to encode the relationships between features corresponding to different regions of an image. In particular, the attention maps from tokens to the class (CLS) token offer a measure of saliency which can be exploited in feature stylization. Based on this idea, we extend our proposed TFS-ViT to have an Attention-aware Token-Level Feature Stylization (ATFS-ViT). 

To this end, we compute the mean of attention matrices of the CLS token over the different attention heads:
\begin{equation}
    \begin{split}
        A (Q, K)_{h, s, s} & \, = \, \text{Softmax}\left( \frac{QK^T}{\sqrt{d_k}} \right)\\ 
        M_{s,s} & \, = \, \frac{1}{H} \sum_{h=1}^{H} A_{h,s,s}\\
        M_{\mr{cls}} & \, = \, M [  0, 1\!:\!S  ]
    \end{split}
\end{equation}
Here, $Q$ and $K$ are the backbone's Query and Key, $H$ is the number of attention heads, and $S$ is the length of the token sequence. In TFS-ViT, we swapped $D$ tokens with their stylized counterparts, the number of which is determined by hyper-parameter $d$. For ATFS-ViT, instead of choosing the tokens randomly, we select those with highest activation in $M_{\mr{cls}}$. The rationale behind this strategy is that, by picking the most active tokens with respect to the CLS token, our method will focus on stylizing the image's foreground, which is more important than the background for the final prediction. Experimental evidence supporting this approach can be found in Sec~\ref{subsection:Further_Analysis}.

\begin{table*}[t]
    \centering
        \caption{\small 
        Comparison to the state-of-art on five benchmarks, reporting the mean and standard deviation across three runs.The best and second best results are in \textbf{bold} and \underline{underlined} fonts, respectively.}
    \label{tab:sota}
    \small
    \tabcolsep=0.05cm
    \adjustbox{max width=\textwidth}{
    \begin{tabular}{lp{0.1cm}cp{0.1cm}cp{0.1cm}cp{0.1cm}cp{0.1cm}cp{0.1cm}cp{0.1cm}cp{0.1cm}c}
\toprule
\textbf{Method}           && \textbf{Backbone}            && \textbf{\#\,Params}            && \textbf{VLCS}   && \textbf{PACS}             && \textbf{OfficeHome}       && \textbf{TerraInc}   && \textbf{DomainNet}        && \textbf{Average}              \\
\midrule
ERM~\cite{Gulrajani2021InSO}          && ResNet-50 &&  23.5M    && 77.5\ppm0.4            && 85.5\ppm0.2            && 66.5\ppm0.3            && 46.1\ppm1.8            && 40.9\ppm0.1            && 63.3                      \\
IRM~\cite{arjovsky2019invariant}               && ResNet-50 &&  23.5M     && 78.5\ppm0.5            && 83.5\ppm0.8            && 64.3\ppm2.2            && 47.6\ppm0.8            && 33.9\ppm2.8            && 61.5                       \\
GroupDRO~\cite{sagawa2019distributionally}          && ResNet-50 &&  23.5M       && 76.7\ppm0.6            && 84.4\ppm0.8            && 66.0\ppm0.7            && 43.2\ppm1.1            && 33.3\ppm0.2            &&   60.7                   \\
Mixup~\cite{yan2020improve}              && ResNet-50 &&  23.5M      && 77.4\ppm0.6            && 84.6\ppm0.6            && 68.1\ppm0.3            && 47.9\ppm0.8            && 39.2\ppm0.1            &&  63.4                     \\
MLDG~\cite{li2018learning}              && ResNet-50 &&  23.5M       && 77.2\ppm0.4            && 84.9\ppm1.0            && 66.8\ppm0.6            && 47.7\ppm0.9            && 41.2\ppm0.1            &&       63.5                \\
CORAL~\cite{sun2016deep}            && ResNet-50 &&  23.5M        && 78.8\ppm0.6            && 86.2\ppm0.3            && 68.7\ppm0.3            && 47.6\ppm1.0            && 41.5\ppm0.1            &&   64.5                    \\
MMD~\cite{li2018domain}             && ResNet-50 &&  23.5M         && 77.5\ppm0.9            && 84.6\ppm0.5            && 66.3\ppm0.1            && 42.2\ppm1.6            && 23.4\ppm9.5            && 58.8                     \\
DANN~\cite{ganin2016domain}          && ResNet-50 &&  23.5M           && 78.6\ppm0.4            && 83.6\ppm0.4            && 65.9\ppm0.6            && 46.7\ppm0.5            && 38.3\ppm0.1            &&     62.6                  \\
CDANN~\cite{li2018deep}          && ResNet-50 &&  23.5M           && 77.5\ppm0.1            && 82.6\ppm0.9            && 65.8\ppm1.3            && 45.8\ppm1.6            && 38.3\ppm0.3            &&       62.0                \\
MTL~\cite{blanchard2017domain}              && ResNet-50 &&  23.5M         && 77.2\ppm0.4            && 84.6\ppm0.5            && 66.4\ppm0.5            && 45.6\ppm1.2            && 40.6\ppm0.1            &&    62.8                   \\
SagNet~\cite{Nam_2021_CVPR}   && ResNet-50 &&  23.5M                && 77.8\ppm0.5            && 86.3\ppm0.2            && 68.1\ppm0.1            && 48.6\ppm1.0            && 40.3\ppm0.1            &&        64.2               \\
ARM~\cite{zhang2021adaptive}              && ResNet-50 &&  23.5M        && 77.6\ppm0.3            && 85.1\ppm0.4            && 64.8\ppm0.3            && 45.5\ppm0.3            && 35.5\ppm0.2            &&  61.7                     \\
VREx~\cite{krueger2021out}             && ResNet-50 &&  23.5M         && 78.3\ppm0.2            && 84.9\ppm0.6            && 66.4\ppm0.6            && 46.4\ppm0.6            && 33.6\ppm2.9            &&        61.9               \\
RSC~\cite{huang2020self}           && ResNet-50 &&  23.5M           && 77.1\ppm0.5            && 85.2\ppm0.9            && 65.5\ppm0.9            && 46.6\ppm1.0            && 38.9\ppm0.5            &&     62.6                  \\
SelfReg~\cite{Kim_2021_ICCV} && ResNet-50 && 23.5M && 77.5\ppm0.0 && 86.5\ppm0.3&& 69.4\ppm0.2 && 51.0\ppm0.4 && 44.6\ppm0.1 && 65.8\\
mDSDI~\cite{bui2021exploiting} && ResNet-50 && 23.5M && 79.0\ppm0.3 &&  86.2\ppm0.2 && 69.2\ppm0.4 && 48.1\ppm1.4 && 42.8\ppm0.1 && 65.0\\ 
SWAD~\cite{cha2021swad} && ResNet-50 && 23.5M && 79.1\ppm0.1 && 88.1\ppm0.1 && 70.6\ppm0.2 && 50.0\ppm0.3 && 46.5\ppm0.1 && 66.8\\ 
\midrule
ERM-ViT~\cite{touvron2021training} && DeiT-Small && 22M && 78.8\ppm0.5 && 84.9\ppm0.9 && 71.4\ppm0.1 &&  43.4\ppm0.5 && 45.5\ppm0.0&& 64.8\\
SDViT~\cite{sultana2022self} && DeiT-Small && 22M && 78.9\ppm0.4  &&  86.3\ppm0.2   &&  71.5\ppm0.2 && 44.3\ppm1.0 && 45.8\ppm0.0 && 65.3 \\ 
 \bf TFS-ViT (ours) && DeiT-Small && 22M && 80.19\ppm0.45 && 87.27\ppm0.38 && 72.08\ppm0.13 && 48.60\ppm0.61 && 46.60\ppm0.06 && 66.95  \\ 
 \bf ATFS-ViT (ours)  && DeiT-Small && 22M &&  \underline{80.65\ppm0.36} && 87.54\ppm0.39 && 71.44\ppm0.16 && 46.06\ppm0.70 && 46.18\ppm0.07 && 66.37  \\ 
 \midrule
ERM-ViT~\cite{yuan2021tokens} && T2T-ViT-14 && 21.5M && 78.9\ppm0.3 && 86.8\ppm0.4  &&  73.7\ppm0.2 && 48.1\ppm0.2 && 48.1\ppm0.1 && 67.1\\ 
SDViT~\cite{sultana2022self} && T2T-ViT-14 && 21.5M && 79.5\ppm0.8   &&  88.0\ppm0.7 &&   74.2\ppm 0.3   && 50.6\ppm0.8 &&   \underline{48.2\ppm0.2} && 68.1 \\
 \bf TFS-ViT (ours)        && T2T-ViT-14 && 21.5M && 80.03\ppm0.25 && \underline{88.99\ppm0.45} && \underline{74.59\ppm0.21} && \textbf{51.76\ppm0.54} && \textbf{48.34\ppm0.13} && \underline{68.74}  \\ 
 \bf ATFS-ViT (ours)  && T2T-ViT-14 && 21.5M && \textbf{80.98\ppm0.40} && \textbf{89.56\ppm0.41} && \textbf{74.65\ppm0.24} && \underline{51.20\ppm0.43} && 47.94\ppm0.21  && \textbf{68.87}  \\ 
\bottomrule

\end{tabular}}
\end{table*}

\section{Experimental Setup}
\label{sec:exp}
\subsection{Datasets}

Following the work of Gulrajani and Lopez-Paz~\cite{Gulrajani2021InSO}, we  compare our approach's performance to the current state-of-art using five challenging datasets, \texttt{PACS}~\cite{li2017deeper}, \texttt{VLCS}~\cite{fang2013unbiased}, \texttt{OfficeHome}~\cite{venkateswara2017deep}, \texttt{TerraIncognita}~\cite{beery2018recognition} and \texttt{DomainNet}~\cite{peng2019moment}, which we describe below.

\texttt{PACS}~\cite{li2017deeper} has a total of 9,991 photos organized into four distinct domains, $d\!\in$\,\{Art, Cartoons, Photos, Sketches\}, and seven distinct classes. \texttt{VLCS}~\cite{fang2013unbiased} is comprised of four different domains, $d\!\in$\,\{Caltech101, LabelMe, SUN09, VOC2007\}, five different classes, and 10,729 different photos. \texttt{OfficeHome}~\cite{venkateswara2017deep} includes four domains, $d\!\in$\,\{Art, Clipart, Product, Real\}, 65 classes, and a total of 15,588 photos. \texttt{TerraIncognita}~\cite{beery2018recognition} contains four camera-trap domains, with 10 categories and a total of 24,778 images. Finally, DomainNet has 6 domains, $d\!\in$\,\{Clipart, Infograph, Painting, Quickdraw, Real, Sketch\}, 345 classes and 586,575 photos.

\subsection{Implementation}
\label{sec:imp}
To have a fair comparison, we implement our method using DomainBed~\cite{Gulrajani2021InSO} -- a recently introduced framework that contains the main existing DG methods and is developed to offer comparisons under a fair evaluation protocol. Accordingly, we follow the same leave-out-one-domain strategy to evaluate performance for different DG datasets, where one domain is used for testing, and the remaining ones are employed for training. Additionally, to choose the best model, 20\% of the training data is used as the validation set\footnote{During training, the model that maximizes the accuracy on this overall validation set is chosen as the best model. The best model is then evaluated on the test domain to report the out-of-domain (unseen domain) accuracy.}. The final result corresponding to each dataset is the average of the accuracy values calculated when using different domains of that dataset as the test domain. To obtain statistically meaningful results, we repeat each experiment three times with different seeds.

Similar to~\cite{sultana2022self}, for all of our experiments, we employ the AdamW optimizer and use the default hyperparameters of DomainBed, including a batch size of $32$, a learning rate of $5e$-$05$, and a weight decay of $0.0$. Additionally, to select the best values of our method-specific hyperparameters, $d$, $n$, we perform a grid search with $d\!\in\!\{0.1, 0.3, 0.5, 0.8\}$ and $n\!\in\!\{1, 2, 3, 4\}$ using the validation set. Most existing methods for DG incorporate ResNet50 as backbone in their architecture. To have a fair comparison, we explore two different ViT-based backbones in our experiments: DeiT~\cite{touvron2021training} and T2T-ViT~\cite{yuan2021tokens}. Specifically, we use DeiT-Small, containing 22M parameters, and T2T-ViT-14, containing 21.5M parameters, since they have a number of parameters comparable to ResNet50 which has 23.5M parameters.

\section{Results}
\label{sec:res}

We first compare the performance of our proposed methods against state-of-art DG methods, across five DG benchmarks. We then present a detailed analysis investigating several aspects of our methods, including the impact of method-specific hyperparameters, the efficacy of various token selection strategies, our proposed method performance in single-source domain generalization, its capability for in-domain regularization, and the computational overhead involved.

\subsection{Comparison with the state-of-the-art}
\label{subsection:Comparison}

In Table~\ref{tab:sota}, we provide a comparison between our TFS-ViT method and 18 recent algorithms for DG implemented in the DomainBed framework~\cite{Gulrajani2021InSO}. We also compare our results with vanilla ERM on the same ViT (ERM-ViT) as well as against SDViT~\cite{sultana2022self} which is currently the only ViT-based method for DG. Using DeiT-Small as backbone, our method improves over ERM-ViT by 2.64\% in PACS, 1.85\% in VLCS, 0.68\% in OfficeHome, 5.2\% in TerraIncognita, and 1.1\% in DomainNet. Moreover, it outperforms the recent SDViT on the same backbone by 1.24\% in PACS, 1.75\% in VLCS, 0.58\% in OfficeHome, 4.3\% in TerraIncognita, and 0.80\% in DomainNet. 

As can be seen, an even greater improvement is achieved when switching the backbone to T2T-ViT. Specifically, TFS-ViT then increases the baseline accuracy by 2.76\% in PACS, 2.08\% in VLCS, 0.95\% in OfficeHome, 3.66\% in TerraIncognita, and 0.24\% in DomainNet. Compared to SDViT on this backbone, our method yields accuracy improvements  of 1.56\% in PACS, 1.48\% in VLCS, 0.45\% in OfficeHome, 1.16\% in TerraIncognita, and 0.14\% in DomainNet. 

On the PACS dataset, our TFS-ViT method achieves a 4.6\% higher accuracy than the vanilla ERM baseline with ResNet-50 backbone, and improved the previous state-of-art by 1.46\%, which was previously held by SWAD~\cite{cha2021swad} with a 88.1\% accuracy. Likewise, we observe a 3.48\% improvement compared to vanilla ERM on the VLCS dataset. Once again, TFS-ViT outperformed the previous state-of-art by a 1.48\% margin, previously held by SDViT~\cite{sultana2022self} with a 79.5\% accuracy. By achieving an accuracy of 75.65\%, our method also improves by 0.45\% the previous record of 74.2\% on the OfficeHome dataset, established by SDViT~\cite{sultana2022self}. For this dataset, a large improvement of 8.15\% over the ERM baseline is achieved by our method. A similar result is observed for the TerraIncognita dataset, for which we witness an increase of 5.66\% over the vanilla ERM, and where TFS-ViT beats the previous record of SDViT~\cite{sultana2022self} by a 0.76\% margin. In DomainNet, we see an improvement of {7.44\%} compared to the ERM baseline. For this last dataset, our method's accuracy of {48.34\%} outperforms the previous record of SDViT~\cite{sultana2022self} by {0.14\%}.

\begin{figure*}[t]
  \centering
  \includegraphics[height=.35\linewidth]{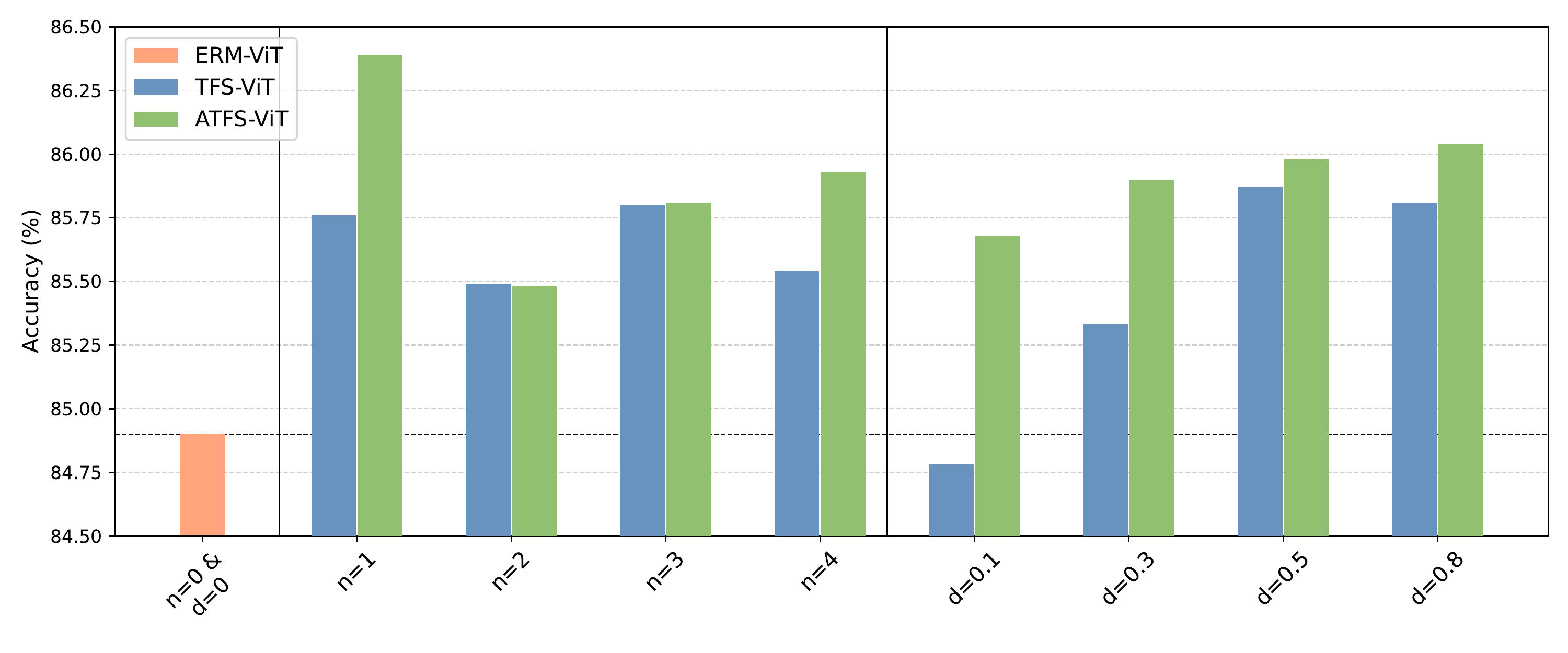}
  \caption{Effects of varying hyperparameters on the PACS dataset. The figure shows the influence of $n$, the number of layers where stylization is performed, with results averaged over different $d$ values, alongside the impact of $d$, the fraction of tokens to be replaced with their stylized counterparts, averaged over various $n$ values.}
  \label{fig:hp_effect}
\end{figure*}

\begin{figure}[t]
  \centering 
  \includegraphics[height=.45\linewidth]{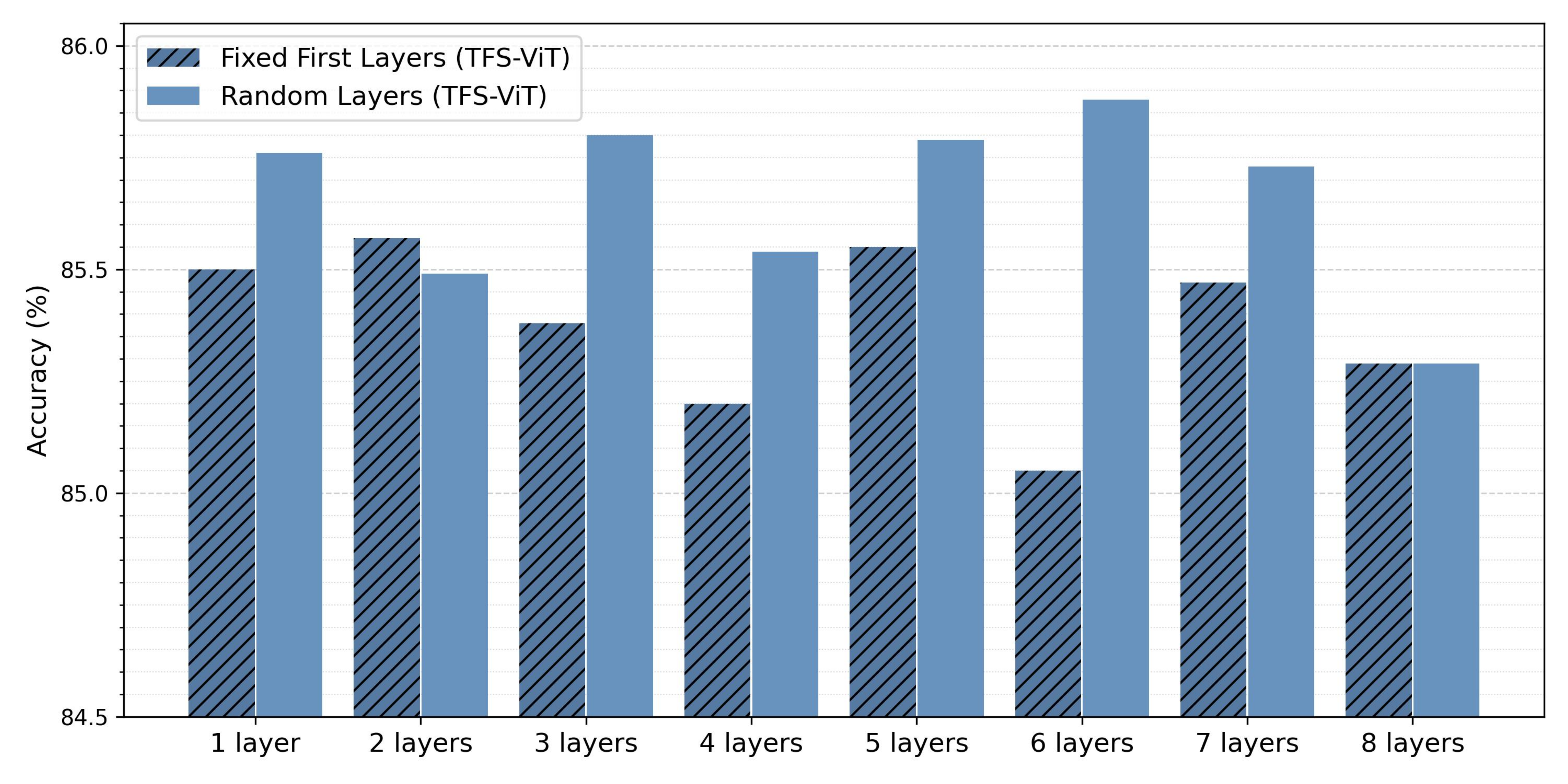}
  \caption{Performance comparison of stylization applied to a fixed initial set of layers versus random layer selection on the PACS dataset. Results are averaged over the different $d$ values.}
  \label{fig:fix_rand}
\end{figure}

\subsection{Further Analyses}
\label{subsection:Further_Analysis} 

\subsubsection{Hyperparameter impact}

Domain Generalization (DG) is an inherently challenging task given the unpredictable nature of unseen domains. For a comprehensive understanding, and to ensure robust outcomes, it is essential to delve into the effects of the method-specific hyperparameters introduced in our paper. As discussed in the Sec~\ref{sec:imp}, our TFS-ViT and ATFS-ViT methods introduce two specific hyperparameters: $n$, which represents the number of random layers where stylization is applied, and $d$, the fraction of tokens to be replaced with their stylized counterparts. TFS-ViT chooses $d$ tokens randomly for stylization, whereas ATFS-ViT selects $d$ tokens with highest activations in $M_{\mr{cls}}$. For all experiments, we conducted a grid search to determine the optimal values for each domain or dataset using the validation set. The grid search parameters were selected in $d\!\in\!\{0.1, 0.3, 0.5, 0.8\}$ and $n\!\in\!\{1, 2, 3, 4\}$ unless otherwise specified.

It is important to note that there is no ``one-size-fits-all'' solution for hyperparameters across different datasets and domains. The optimal choice often depends on the specific characteristics of the dataset and the degree of domain shift. To provide a clearer understanding, we performed an extensive study on the effects of these hyperparameters on the proposed TFS-ViT and ATFS-ViT. The resulting insights of varying these hyperparameters, $n$ and $d$, on the PACS dataset are presented in Fig.~\ref{fig:hp_effect}.

As can be seen, the majority of hyperparameter combinations yield enhanced generalization when compared to the baseline, ERM-ViT. The best hyperparameters are around $n$=3 and $d$=0.5 for the PACS dataset. It is worth mentioning that this figure reports average results across various domains. However, when the domain shift is more severe, we find that higher values of $d$ and $n$ are needed, and vice versa. For example, in the PACS dataset, $n$=2 and $d$=0.1 are the ideal hyperparameters when target domain is Cartoon, which contains less domain shift compared to source domains. On the other hand, when the Sketch domain presenting a more severe domain shift stands as the target, $n$=4 and $d$=0.8 emerges as the most effective combination, aligning with our expectations. Additionally, we can see that the performance of ATFS-ViT is more robust than TFS-ViT in all cases, as it chooses the candidate tokens for stylization in a more intelligent way.

\subsubsection{Fixed Layers vs Random Layers}
%

The use of CNNs for DG relies on several principles, one of which being that information pertaining to style is encoded in the first few layers. As we move closer to the classification head, more information related to classes is included in the feature maps. As a result, CNN-based feature stylization techniques focus on the first few layers of the network for augmenting features. However, the same strategy may not be optimal for ViTs, where features encoding structure can be found in all layers. 

To explore this question, we compare in Figure~\ref{fig:fix_rand} the accuracy on the PACS dataset while conducting feature stylization on the first $n$ layers \emph{vs} randomly selecting $n$ layers from the first $75\%$ of the layers (i.e., the first 8 layers of DeiT). As can be seen, applying stylization to randomly selected layers is usually better than in the first ones. The improved performance achieved by randomly selected layers is also due to the added stochasticity, which exposes the model to a broader range of domain shifts.

\begin{table}[t]
    \centering
    \caption{\small Performance comparison of token selection strategies for stylization across the PACS dataset domains. The best results are highlighted in \textbf{bold}. Descriptions: \textbf{All} - select and stylize all tokens; \textbf{Random} - random selection of tokens for stylization; \textbf{High} and \textbf{Low} - selection based on the highest and lowest activations in $M_{\mr{cls}}$ for stylization, respectively.}
    \label{tab:token_activation}
    \small
    \tabcolsep=0.05cm
    \adjustbox{max width=0.65\textwidth}{
    \begin{tabular}{lp{0.1cm}cp{0.1cm}cp{0.1cm}cp{0.1cm}cp{0.1cm}c}
    \toprule
    \textbf{Token Selection} && \textbf{Art} && \textbf{Cartoon} && \textbf{Photo} && \textbf{Sketch} && \textbf{Average} \\
    \midrule
    Baseline (ERM-ViT) && 87.4&& 81.5 && 98.1 && 72.6 && 84.9\  \\
    \midrule
    \bf All && 88.88 && 82.20 && 98.40  && 76.73  && 86.43 \\
    \bf Random (TFS-ViT) && 89.63 && \textbf{83.03} && \textbf{98.58}  && 77.83  && 87.27 \\
    \bf Low && \textbf{90.46} && 82.93 && 98.39 && 77.39 && 87.29 \\
    \bf High (ATFS-ViT) && \textbf{90.46} && 83.00 && 98.43 && \textbf{78.25}  && \textbf{87.54}  \\
    \bottomrule
    \end{tabular}
    }
\end{table}

\subsubsection{Token Selection Choices}

While our proposed TFS-ViT method has demonstrated impressive accuracy in various domain generalization scenarios, the strategy for token selection, particularly the one based on CLS attention scores, can further improve the performance. In this section, we present a comprehensive analysis to understand the impact of various selection strategies on model performance.

Table~\ref{tab:token_activation} showcases the comparative performance of diverse token selection strategies applied to the PACS dataset. Evidently, the ATFS-ViT method, which stylizes tokens with the highest activations in the CLS attention maps, outperforms not only the TFS-ViT approach that employs random selection, but also strategies that stylizes the lowest activations in CLS attention maps or involve stylizing all tokens (denoted as ``All'' in the table). ``All'' is analogous to TFS-ViT when $d=1$, and can be seen as an adaptation of the MixStyle method~\cite{zhouMixStyleNeuralNetworks2021} into ViTs, a technique originally designed for CNNs. While this approach does exhibit improvement over the baseline, our proposed methods, which introduce greater diversity to the model (please see the Fig~\ref{fig:stylization}), offer a substantial performance boost.

The results of the table suggests that tokens with higher activations in CLS attention maps contain more discriminative features. When stylized, they contribute more effectively to the domain generalization task, compared to those with lower activations which might be less informative or potentially noisy. Furthermore, our observations indicate that using a more powerful ViT backbone (for example T2T-ViT-14) significantly amplifies the performance benefits of the high activation token selection strategy. This observation is evident from the main results in Table~\ref{tab:sota}.

\subsubsection{Single Source Domain Generalization}

       
    
\begin{figure}[t]
  \centering
   \includegraphics[height=.6\linewidth]{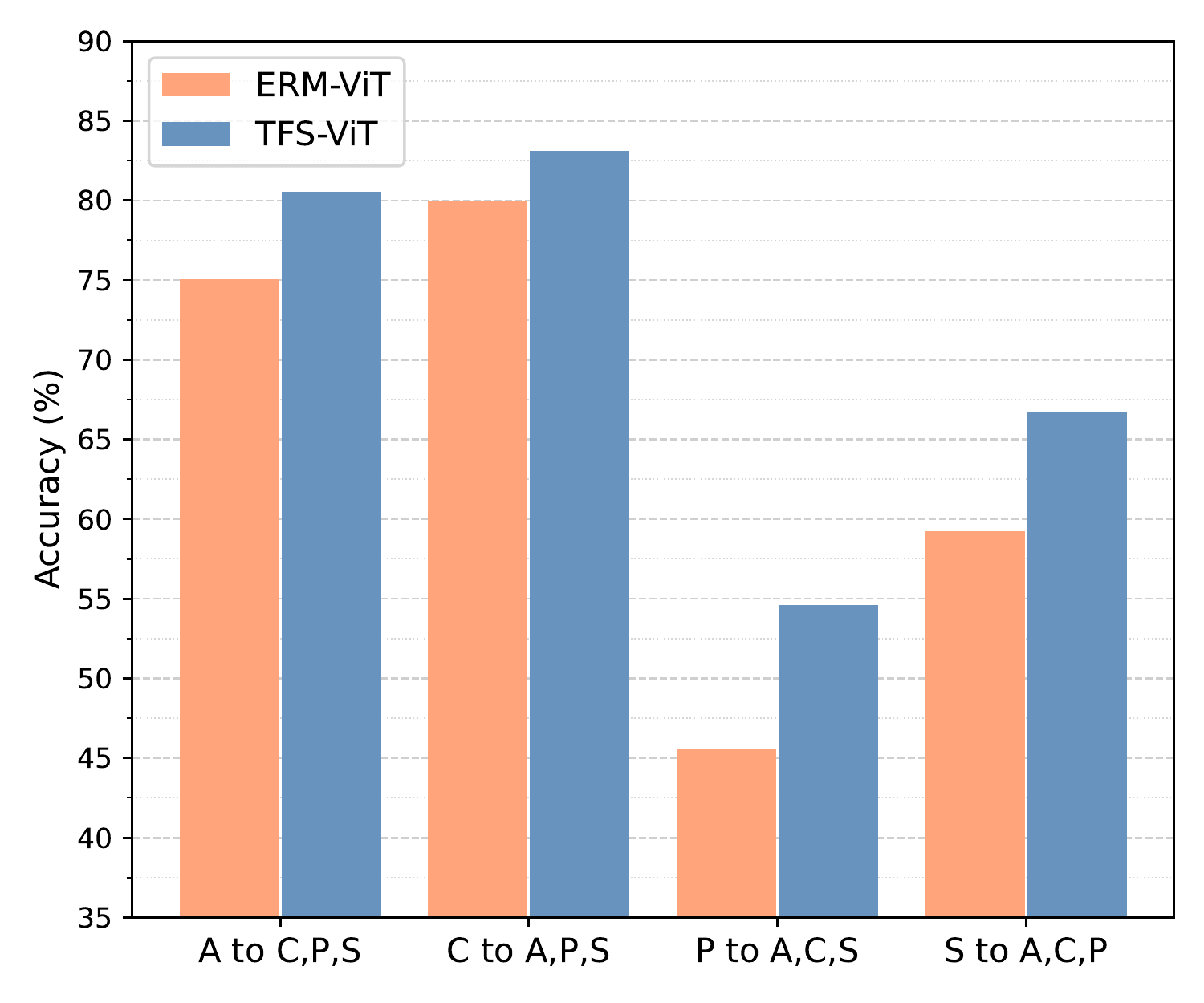}
   \caption{Comparison of ERM-ViT and TFS-ViT performance in Single-Source Domain Generalization setting on the PACS dataset.}
   \label{fig:ssdg}
\end{figure}

While it is generally assumed that all samples from a given domain originate from the same distribution, this assumption may not always hold in practice. Based on this idea, we evaluate the advantage of using our TFS-ViT method when training with images from the same domain. Figure~\ref{fig:ssdg} compares the performance of ERM-ViT and our TFS-ViT method, when training with a single source domain of the PACS dataset and testing on all others. As shown, TFS-ViT considerably increases the generalization capability of ERM-ViT for every source domain.

\subsubsection{Regularization Effect}

\begin{figure}[t]
  \centering
   \includegraphics[height=.6\linewidth]{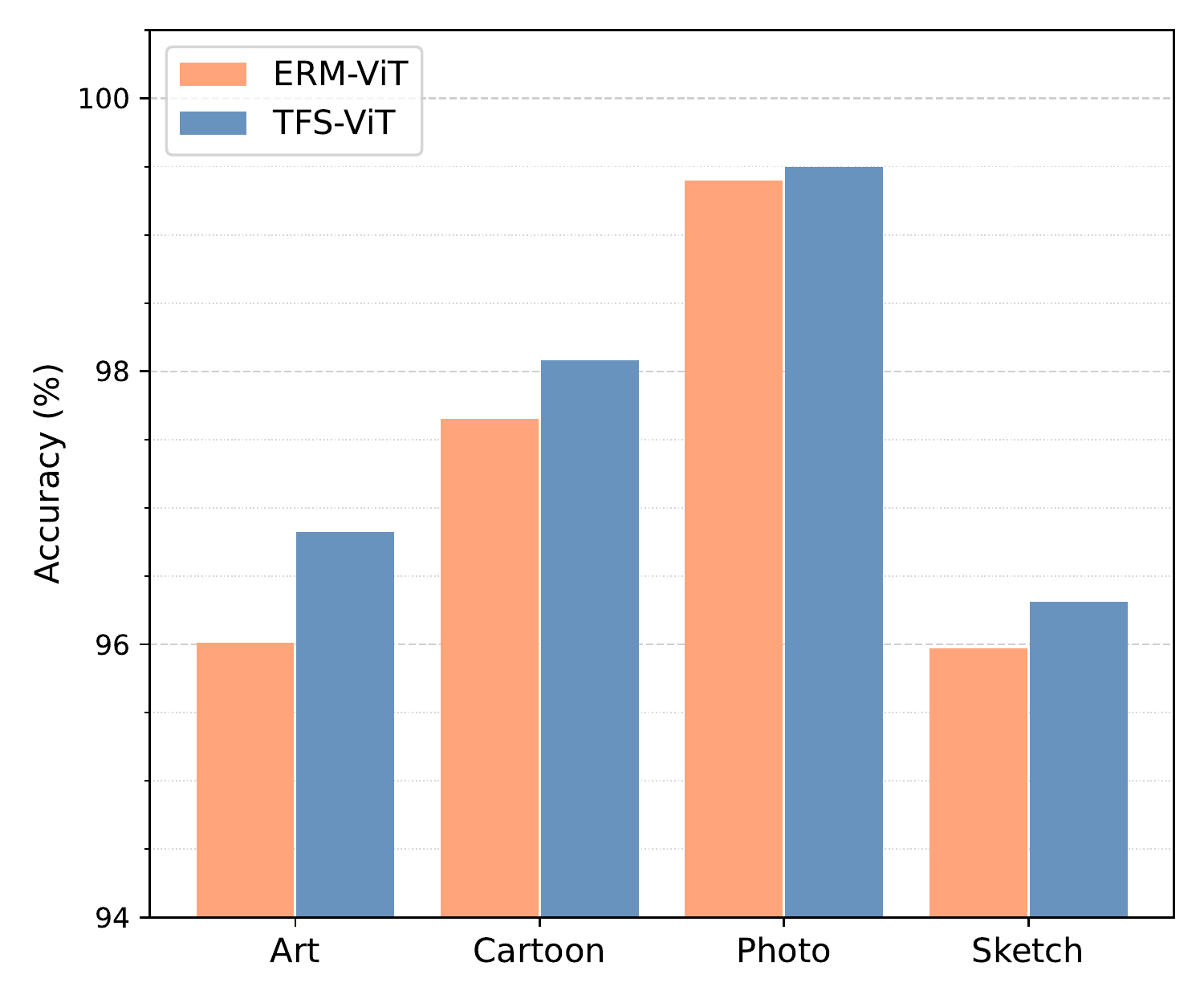}
   \caption{Regularization effect. Comparison of ERM-ViT and TFS-ViT performance when training and evaluation is done on the same domain for different domains of the PACS dataset.}
   \label{fig:reg}
\end{figure}

In the next analysis, we evaluate whether supplementing the network with synthetic features generated by TFS-ViT can also improve accuracy when evaluating the network on the same domain it was trained on. For this analysis, we train and test the model separately on each domain of the PACS dataset. Results presented in Figure~\ref{fig:reg} reveal that TFS-ViT also achieves a significantly higher accuracy than ERM-ViT in this setting. This demonstrates the usefulness of TFS-ViT in a standard in-domain setting, in addition to the OOD scenario of DG. Therefore, our  method can be also regarded as a regularization technique, and can be employed across a variety of applications.

\subsubsection{Detailed Results on the PACS Dataset}

\begin{table*}[t]
\begin{center}
\caption{Our proposed method performance on different domains of the PACS~\cite{li2017deeper} dataset. Mean and Standard Deviation are reported across three runs. The best and second best average is in \textbf{bold} and \underline{underlined} fonts, respectively.}
\label{tab:ablation_PACS}
\adjustbox{max width=\textwidth}{
\begin{tabular}{lccccccc}
\hline \noalign{\smallskip}
\textbf{Method} & \textbf{Backbone} & \textbf{\#\,of Params} &  \textbf{Art} & \textbf{Cartoon} & \textbf{Photos} & \textbf{Sketch} & \textbf{Average} \\
\noalign{\smallskip}
\toprule
\noalign{\smallskip}
ERM  & ResNet-50 & 23.5M & 81.3\ppm0.6 & 80.9\ppm0.3& 96.3\ppm0.6  &  78.0\ppm1.6 & 84.1\ppm0.4  \\ 
\midrule
ERM-ViT  & DeiT-Small & 22M &  87.4\ppm1.2 &81.5\ppm0.8&	98.1\ppm0.1	& 72.6\ppm3.3&	84.9\ppm0.9   \\
SDViT~\cite{sultana2022self} & DeiT-Small & 22M & 87.6\ppm0.3   &       82.4\ppm0.4      &    98.0\ppm0.3     &     77.2\ppm1.0 & 86.3\ppm0.2     \\
\bf TFS-ViT (ours) & DeiT-Small & 22M &  89.63\ppm 0.86   &   83.03\ppm0.31     &  98.58  \ppm 0.19   &  77.83   \ppm1.21 & 87.27\ppm0.38     \\
\bf ATFS-ViT (ours) & DeiT-Small & 22M &  90.46\ppm0.67    &  83.00\ppm0.31     &   98.43\ppm 0.15	 &  78.25  \ppm1.37	 & 87.54\ppm0.39    \\
\midrule
ERM-ViT  & T2T-ViT-14 &  21.5M & 89.6\ppm0.9 & 81.0\ppm0.9 &  \textbf{98.9\ppm0.2}    &  77.6\ppm2.6   & 86.8\ppm0.4    \\
SDViT\cite{sultana2022self} & T2T-ViT-14 & 21.5M & 90.2\ppm1.2  &        82.7\ppm0.7   &       98.6\ppm0.2       &   80.5\ppm2.2    &      88.0 \ppm0.7  \\
\bf TFS-ViT (ours) & T2T-ViT-14 & 22M &  \underline{90.48\ppm0.72}    &    \underline{83.62\ppm0.52}   & \underline{98.80\ppm0.21}     &  \underline{83.04\ppm1.56} &   \underline{88.99\ppm0.45}  \\
\bf ATFS-ViT (ours) & T2T-ViT-14 & 22M &  \textbf{90.48\ppm0.15}  &  \textbf{84.86\ppm1.14} &   98.53\ppm0.02	 &  \textbf{84.38\ppm1.18}  	 &  \textbf{89.56\ppm0.41}   \\
\bottomrule
\end{tabular}
}
\end{center}
\end{table*}

The PACS dataset contains highly different domains, ranging from photos to basic sketches. For the following analysis, we use this dataset to evaluate the robustness of TFS-ViT to such domain variability. Toward this goal, we give in Table~\ref{tab:ablation_PACS} a breakdown of our method's performance across all PACS domains. As can be seen, TFS-ViT outperforms vanilla ERM-ViT on all domains but one (Photos), improving the average accuracy by $2.76\%$. In particular, it achieves a significant improvement of $6.78\%$ for the Sketch domain, which is the most challenging one due to its large domain shift.

\subsubsection{Computational Overhead}

\begin{table}[t]
    \centering
        \caption{Computational Statistics for training on three source domains of the PACS dataset for 5000 steps with a batch size of 32.}
    \label{tab:overhead}
    \small
    \adjustbox{max width=\textwidth}{
    \begin{tabular}{lcc}
    \toprule
    \bf Model    & \bf Training Time (hrs) & \bf GPU Mem (GB) \\ 
    \midrule
    ERM-ViT & 0.36769             & 7.02028      \\ 
    \bf TFS-ViT (ours)  & 0.37166             & 7.02428      \\ 
    \bf ATFS-ViT (ours) & 0.37317             & 7.02450      \\ 
    \bottomrule
    \end{tabular}}
\end{table}

The computational overhead of domain generalization approaches compared to vanilla models is a major roadblock to their use in real-world applications. To demonstrate our method's computational efficiency, we compare in Table \ref{tab:overhead} the training times and GPU memory requirements of TFS-ViT against ERM-ViT, for our biggest model which has four layers. As reported, TFS-ViT only incurs a 1.08 percent increase in training time and a 0.06 percent increase in GPU memory. This suggests that TFS-ViT can be employed without having to worry about added computational costs.



\subsubsection{Extendability Analysis}

\begin{figure}[t]
  \centering
   \includegraphics[height=.6\linewidth]{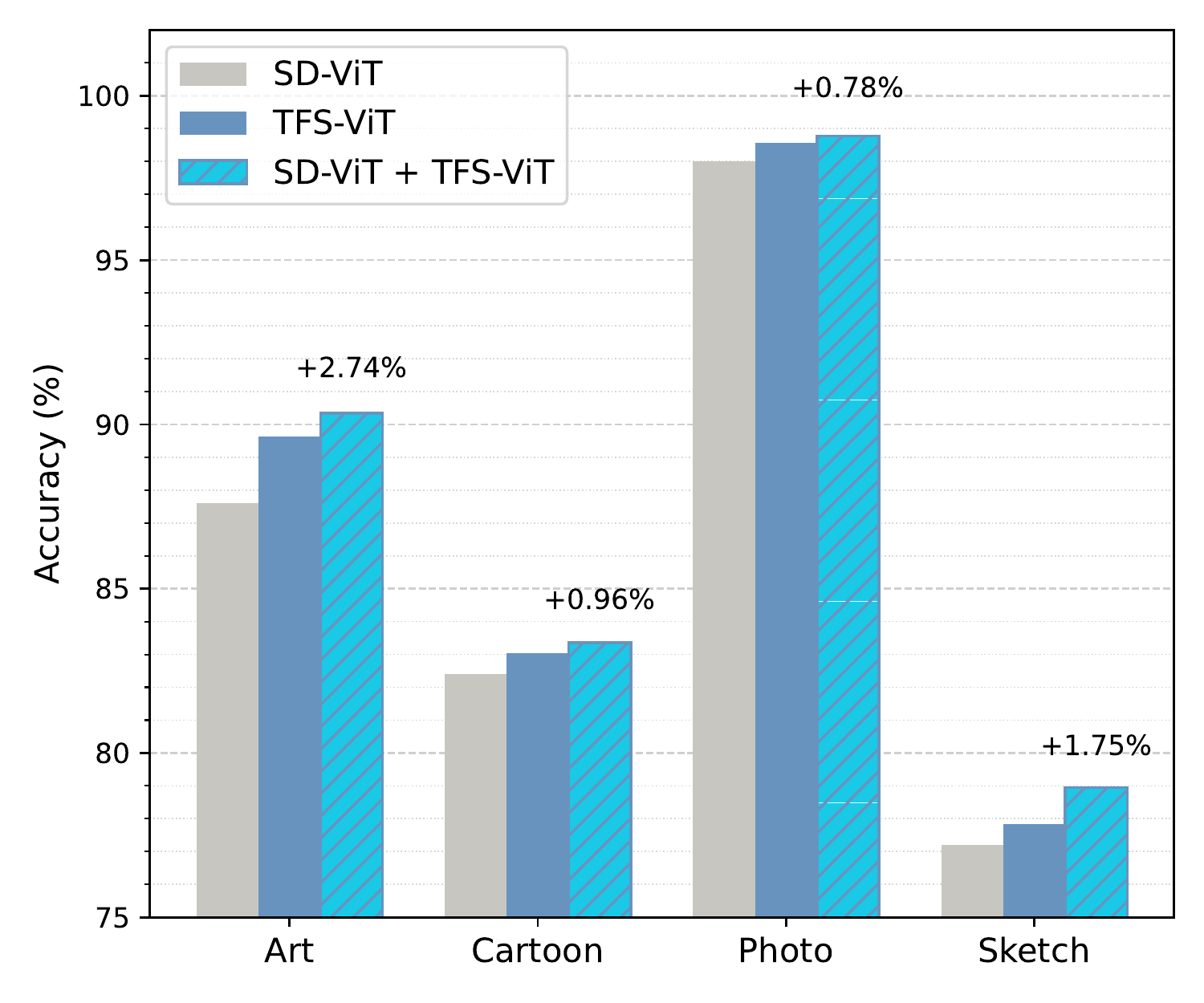}
   \caption{Comparison between the performance of TFS-ViT when it is applied to SDViT and the original SDViT and TFS-ViT on different domains of the PACS dataset. Annotations on the bars indicate the percentage increase in accuracy achieved by SDViT $+$ TFS-ViT, compared to the original SDViT. The results show the extendability of our method which can be applied on top of any ViT-based method with negligible increased computational complexity.}
   \label{fig:sd+tfs}
\end{figure}

Our TFS-ViT method is flexible and, since it has a low computational cost and simply requires to mix inner tokens of the backbone, it can be used as a module on top of any backbone or in tandem with other domain generalization techniques. To show the complementary benefit brought by our method, we added it on top of the Self-Distilled Vision Transformer (SDViT) approach for DG \cite{sultana2022self}, which regularizes training with auxiliary losses in intermediate layers. As shown in Figure~\ref{fig:sd+tfs}, TFS-ViT further improves the performance of SDViT on different domains of the PACS dataset.

\subsubsection{Visualization of Attention Maps}

\begin{figure*}[t]
  \centering
   \includegraphics[width=.97\linewidth]{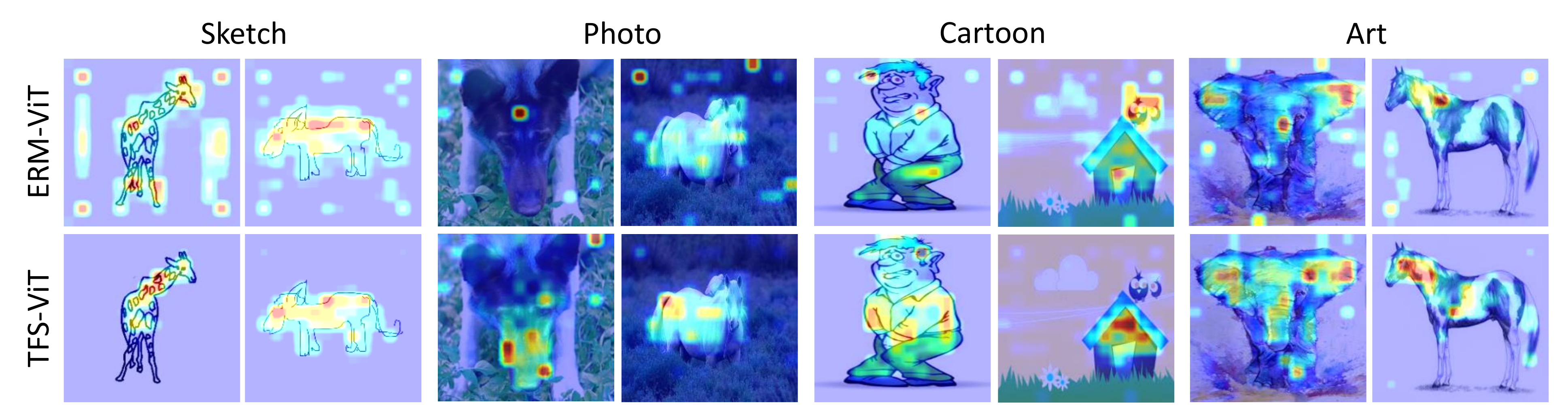}
   \caption{Comparison of attention maps for the CLS token of the last layer generated by two models, ERM-ViT (baseline) and TFS-ViT (with DeiT-Small backbone), on various domains of the PACS dataset as the unseen/target domain.}
   \label{fig:vis}
\end{figure*}

In contrast to CNNs, which learn from local patterns, ViTs attempts to represent global relationships using multi-head self-attention (MSA) layers. As a result, by visualizing the attention maps of CLS token, we may get access to the most decisive parts of the input. Figure~\ref{fig:vis} depicts visual comparisons of final layer attention maps for ERM-ViT and TFS-ViT. It demonstrates that our method assists networks in attending to features that are more indicative of the semantics of the picture in all target domains, which lead to improving the overall generalization performance.

\section{Conclusion}
\label{sec:conclusion}

In this paper, we presented the first token-level feature stylization approach to improve the generalization capabilities of ViTs in out-of-distribution scenarios. We also proposed an innovative attention-aware stylization technique that makes use of attention maps in MSA layers to guide the augmentation toward relevant regions of the image. In a comprehensive set of experiments using five challenging benchmark datasets, we showed our TFS-ViT method to outperform existing alternatives for DG, and to achieve state-of-art accuracy on these datasets. Detailed analyses revealed the benefit of randomly selecting layers on which to perform stylization, as well as its usefulness in single-source domain generalization and in-domain settings. Our method provides consistent improvements for very different domains, ranging from photos to sketches, has a negligible overhead in terms of training time and GPU memory, and can further boost performance when used in conjunction with other DG strategies such as self-distillation.

In this work, we demonstrated the advantage our method on two well-known ViT backbones, DeiT-Small and T2T-ViT-14, which have a number of parameters comparable to the standard ResNet-50 architecture. However, additional improvements could be achieved for more recent ViT architectures, for instance, the Swin transformer \cite{liu2021swin} which uses a shifted window strategy to learn image representations in a hierarchical manner. Moreover, while we exploited the attention maps of class tokens to steer the stylization toward salient regions in the image, more sophisticated techniques could be considered. For instance, future work could investigate the idea of stylizing foreground and background regions separately.

\bibliographystyle{IEEEtran}
\bibliography{egbib}


\end{document}


\title{TFS-ViT: Token-Level Feature Stylization for Domain Generalization Supplementary Material}

\author{\IEEEauthorblockN{\textbf{Mehrdad Noori}, 
\textbf{Milad Cheraghalikhani},
\textbf{Ali Bahri}, 
\textbf{Gustavo A. Vargas Hakim}, \\
\textbf{David Osowiechi}, 
\textbf{Ismail Ben Ayed}, 
\textbf{Christian Desrosiers}} 
\\ \vspace{0.3cm}\IEEEauthorblockA{LIVIA, ÉTS Montreal, Quebec, Canada}
\\ \IEEEauthorblockA{International Laboratory on Learning Systems (ILLS)}
}
\markboth{}{}

\maketitle

\section{Result Reproduction}
Our proposed TFS-ViT method is implemented in Python and PyTorch framework, and we use an NVIDIA Tesla V100 GPU for all of our experiments. The original implementation and the instructions for reproducing the results can be found in \href{https://github.com/Mehrdad-Noori/TFS-ViT_Token-level_Feature_Stylization}{https://github.com/Mehrdad-Noori/TFS-ViT\_Token-level\_Feature\_Stylization}.

\section{Stylization Visualization}
In order to get a better understanding of our proposed token-level feature stylization method, we train a simple ViT-based encoder-decoder network\footnote{We use ViT-base~\cite{dosovitskiy2020image} as encoder, and an architecture similar to~\cite{he2022masked} as our decoder. The "Photo", "Art", "Cartoon" domains of PACS dataset is used to train the model. During the training, no feature stylization is used. We train the network for 800 epochs using the default hyperprameters in~\cite{he2022masked}.} without performing any stylization. When the training is finished, we perform token-level stylization in the encoder using a batch of input images - precisely like what we do in TFS-ViT - and try to reconstruct images to see the effect of stylization at the pixel level. For this section, we conduct two experiments: the effects of single layer selection as well as random selections of multiple layers on the stylization using two values, $\{ 0.5, 1.0\}$, for the parameter $d$.

\begin{figure*}[!b]
  \centering
   \includegraphics[width=1.0\linewidth]{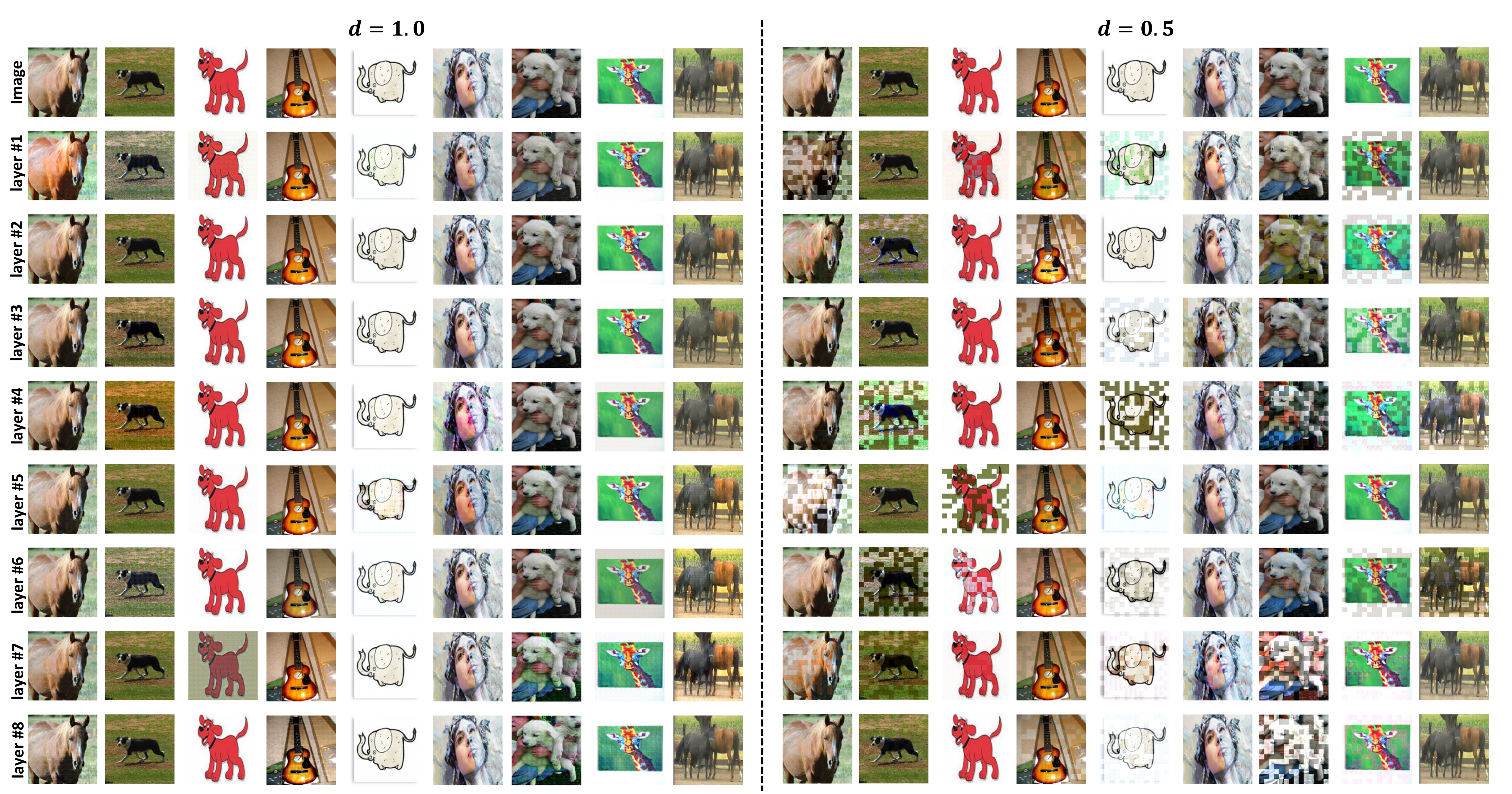}
  \caption{The effect of the token-level stylization on different layers of the Vision Transformer model (encoder). The left and right figures show the results of the reconstructed images when all and half of the tokens are stylized, respectively.}
  \label{fig:per_layer}
\end{figure*}

\textbf{Single Layer} Figure~\ref{fig:per_layer} demonstrates the reconstructed images when token-level stylization is performed on different layers of the Vision Transformer model (encoder). As can be observed from the figure, stylizing tokens at different layers of the model preserves the original structure of the input images. This observation is in contrast with CNNs in which feature-level stylization can only be adopted on the very first convolutional layers where style-related features are extracted~\cite{jeon2021feature, zhou2021domain}. Additionally, from the figure, we observe that choosing a random number of tokens (in this case $d=0.5$) and replacing them with stylized ones leads to creating more diverse samples. 

\begin{figure*}[!b]
  \centering
   \includegraphics[width=1.0\linewidth]{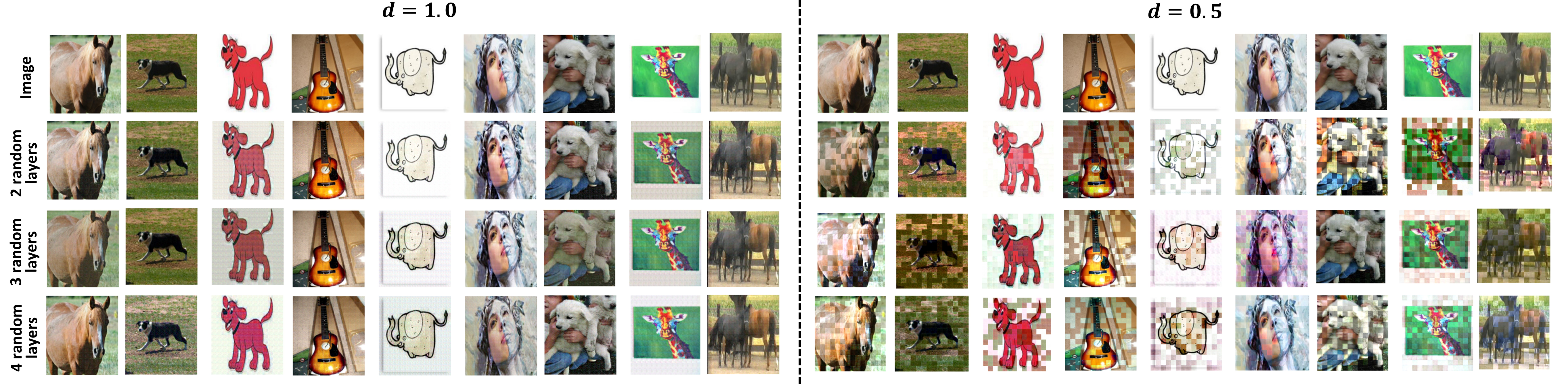}
  \caption{The effect of stylizing multiple randomly-selected layers on the reconstructed images. The left and right figures show the results of the reconstructed images when all and half of the tokens are stylized, respectively.}
  \label{fig:random_layer}
\end{figure*}

\textbf{Multiple Random Layers} Figure~\ref{fig:random_layer} illustrates the effect of choosing up to four random layers on the reconstructed images. As can be seen from the figure, stylizing more layers in addition to stylizing random tokens (in this case $d=0.5$) results in even more diverse images compared to single-layer selection. These diverse synthetic features generated from different domains are able to simulate different kinds of domain shifts during training and accordingly force the network to learn domain-invariant features.

\section{Detailed Results}

The detailed breakdown of our method's performance across all domains for VLCS~\cite{fang2013unbiased}, OfficeHome~\cite{venkateswara2017deep}, TerraIncognita~\cite{beery2018recognition}, and DomainNet~\cite{peng2019moment} can be seen in Tables. \ref{tab:ablation_vlcs}, \ref{tab:ablation_oh}, \ref{tab:ablation_terra}, and \ref{tab:ablation_dn}, respectively. 
As can be seen, our method has improves the average accuracy of baseline (ERM-ViT) and SD-ViT, which is the only ViT-based method for Domain generalisation that is currently available for all these four datasets. This is also true for most of the target domains on these datasets, not just on average, which can demonstrate our method's robustness in dealing with different kinds of domain shift.

\section{Additional Attention Map Visualizations}

\begin{figure*}[t]
  \centering
   \includegraphics[width=.95\linewidth]{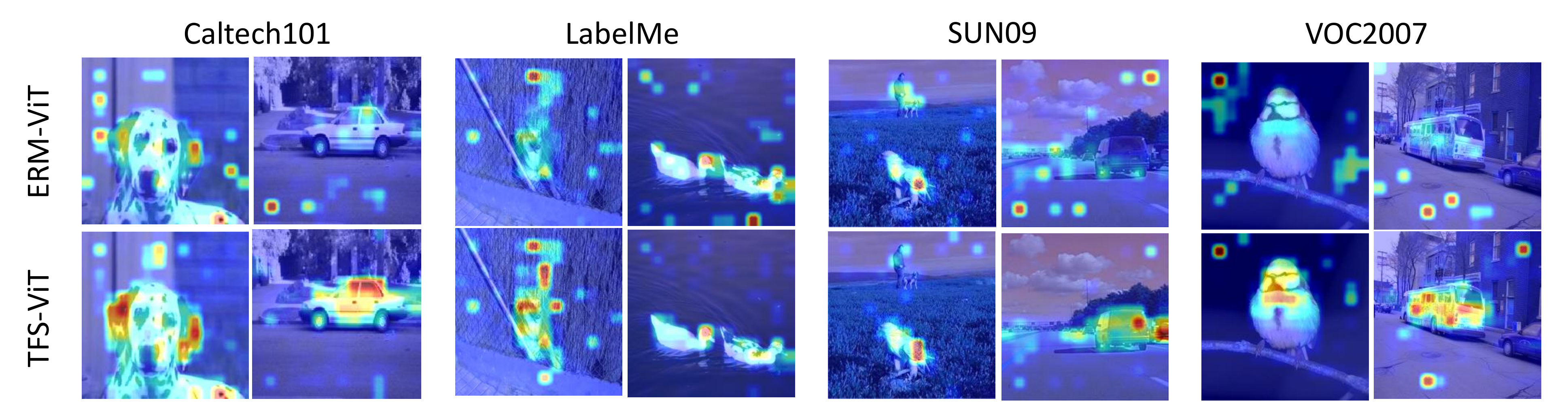}
   \caption{Comparison of attention maps for the CLS token of the last layer generated by two models, ERM-ViT (baseline) and TFS-ViT (with DeiT-Small backbone), on various domains of the VLCS dataset as the unseen/target domain.}
   \label{fig:vis_vlcs}
\end{figure*}

\begin{figure*}[t]
  \centering
   \includegraphics[width=.95\linewidth]{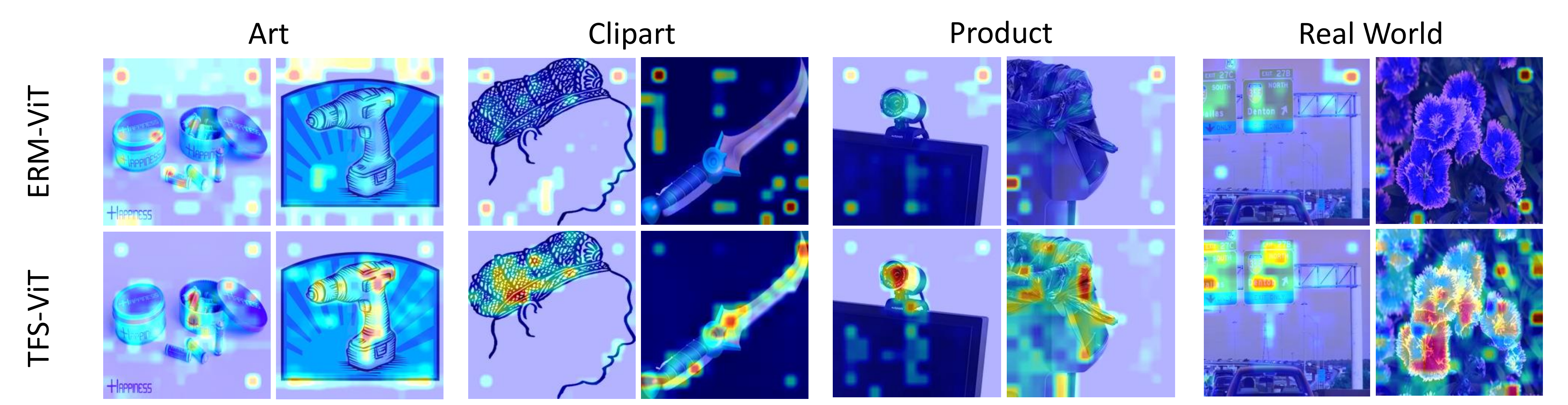}
   \caption{Comparison of attention maps for the CLS token of the last layer generated by two models, ERM-ViT (baseline) and TFS-ViT (with DeiT-Small backbone), on various domains of the OfficeHome dataset as the unseen/target domain.}
   \label{fig:vis_oh}
\end{figure*}

\begin{figure*}[t]
  \centering
   \includegraphics[width=.95\linewidth]{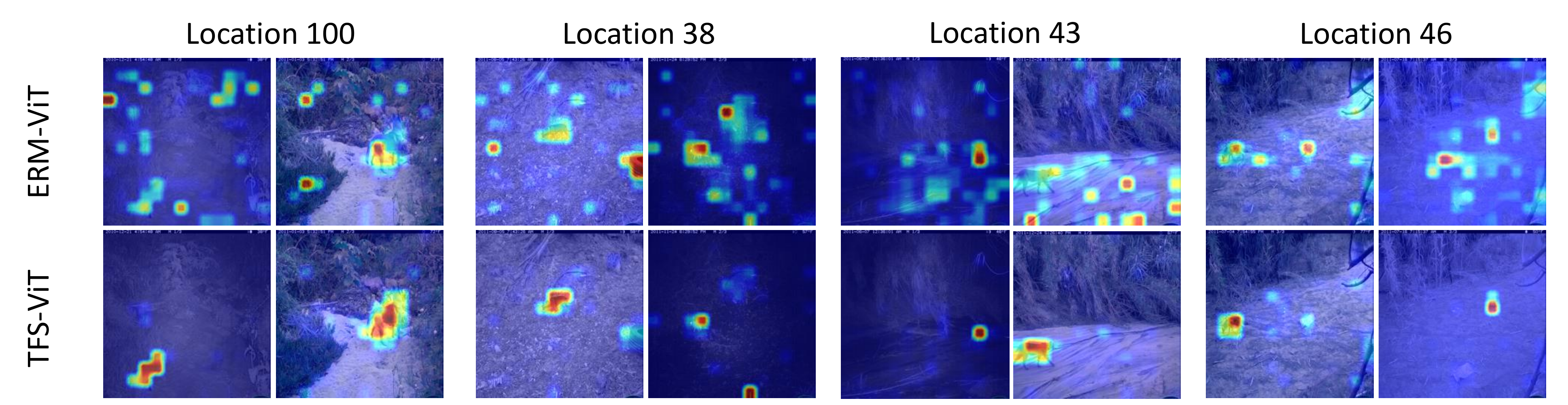}
   \caption{Comparison of attention maps for the CLS token of the last layer generated by two models, ERM-ViT (baseline) and TFS-ViT (with DeiT-Small backbone), on various domains of the TerraIncognita dataset as the unseen/target domain.}

   \label{fig:vis_terra}
\end{figure*}

In Figs. \ref{fig:vis_vlcs}, \ref{fig:vis_oh}, and \ref{fig:vis_terra}, we compared attention maps of CLS Token of our method with baseline for several images from all possible target domains on the VLCS~\cite{fang2013unbiased}, OfficeHome~\cite{venkateswara2017deep}, and TerraIncognita~\cite{beery2018recognition} datasets. In almost all of these scenarios, the cls token of our method mostly uses tokens that represent the foreground object instead of background's token which mostly contains style-related information.But the baseline approach (ERM-ViT) focuses more on the features of the background and less on the features of the object in the foreground.

\bibliographystyle{IEEEtran}
\bibliography{egbib}

\begin{table*}[t]
\begin{center}
\caption{ Our proposed method performance on different domains of the VLCS~\cite{fang2013unbiased} dataset. Mean and Standard Deviation are reported across three runs. The best and second best average is in \textbf{bold} and \underline{underlined} fonts, respectively.}
\label{tab:ablation_vlcs}
\adjustbox{max width=\textwidth}{
\begin{tabular}{lccccccc}
\toprule \noalign{\smallskip}
\textbf{Method} & \textbf{Backbone} & \textbf{\#\,of Params} &  \textbf{Caltech101} & \textbf{LableMe} & \textbf{SUN09} & \textbf{VOC2007} & \textbf{Average} \\
\noalign{\smallskip}
\midrule
\noalign{\smallskip}
ERM-ViT  & DeiT-Small & 22M & 96.7\ppm0.8     &     65.2\ppm1.0      &    73.9\ppm0.3     &     77.4\ppm0.3 & 78.3\ppm0.5  \\
SDViT~\cite{sultana2022self} & DeiT-Small & 22M & 96.8\ppm0.5      &    64.2\ppm0.8   &       76.2\ppm0.4      &    78.5\ppm0.4 & 78.9\ppm0.4     \\
TFS-ViT (ours) & DeiT-Small & 22M &  98.70\ppm0.35   &   \underline{66.57\ppm0.51}     &  76.5\ppm1.38   &  79.00\ppm1.00 & 80.19\ppm0.45     \\
ATFS-ViT (ours) & DeiT-Small & 22M &  \textbf{99.12\ppm0.25}   &  \textbf{66.71\ppm0.77}     &    75.96\ppm0.93	 &  \underline{80.82\ppm0.74} 	 & \underline{80.65\ppm0.36}    \\
\midrule
ERM-ViT  & T2T-ViT-14 &  21.5M & 96.5\ppm0.5 & 64.5\ppm0.1     &     76.4\ppm0.4 &         78.2\ppm1.0 & 78.9\ppm0.3   \\
SDViT\cite{sultana2022self} & T2T-ViT-14 & 21.5M & 96.9\ppm0.4    &  64.0\ppm0.5  &        76.7\ppm1.4   &  80.4\ppm1.3    & 79.5\ppm0.8   \\
TFS-ViT (ours) & T2T-ViT-14 & 21.5M &  98.32\ppm0.29    &    63.86\ppm0.83   & \underline{77.57\ppm0.43}     &  80.37\ppm0.15 &  80.03\ppm0.25  \\
ATFS-ViT (ours) & T2T-ViT-14 & 21.5M &  \underline{98.82\ppm0.13}  &  65.84\ppm1.30 &   \textbf{78.32\ppm0.64}	 &  \textbf{80.92\ppm0.65}  	 &  \textbf{80.98\ppm0.40}   \\
\bottomrule
\end{tabular}
}
\end{center}
\end{table*}

\begin{table*}[t]
\begin{center}
\caption{Our proposed method performance on different domains of the OfficeHome~\cite{venkateswara2017deep} dataset. Mean and Standard Deviation are reported across three runs. The best and second best average is in \textbf{bold} and \underline{underlined} fonts, respectively.}
\label{tab:ablation_oh}
\adjustbox{max width=\textwidth}{
\begin{tabular}{lccccccc}
\toprule \noalign{\smallskip}
\textbf{Method} & \textbf{Backbone} & \textbf{\#\,of Params} &  \textbf{Art} & \textbf{Clipart} & \textbf{Product} & \textbf{Real World} & \textbf{Average} \\
\noalign{\smallskip}
\midrule
\noalign{\smallskip}
ERM-ViT  & DeiT-Small & 22M &  67.6\ppm0.3    &57.0\ppm0.6 &     79.4\ppm0.1    & 81.6\ppm0.4 & 71.4\ppm0.1    \\
SDViT~\cite{sultana2022self} & DeiT-Small & 22M & 68.3\ppm0.8   &  56.3\ppm0.2   & 79.5\ppm0.3  &    81.8\ppm0.1   &  71.5\ppm0.2    \\
TFS-ViT (ours) & DeiT-Small & 22M &  68.52\ppm0.22   &   57.73\ppm0.08    &  80.15\ppm0.30   &  81.90\ppm0.37 & 72.08\ppm0.13     \\
ATFS-ViT (ours) & DeiT-Small & 22M &  67.39\ppm0.20   &  56.98\ppm0.31     &   79.40\ppm0.25	 &   81.97\ppm0.45	 & 71.44\ppm0.16    \\
\midrule
ERM-ViT  & T2T-ViT-14 &  21.5M &  70.2\ppm0.5  &        59.0\ppm0.6      &    81.9\ppm0.3   &       \underline{83.6\ppm0.6} &  73.7\ppm0.2  \\
SDViT\cite{sultana2022self} & T2T-ViT-14 & 21.5M &  71.1\ppm0.5      &    59.2\ppm0.3      &    \textbf{82.8\ppm0.4}      &    83.5\ppm0.3      & 74.2\ppm0.3  \\
TFS-ViT (ours) & T2T-ViT-14 & 21.5M &  \underline{71.37\ppm0.60}    &    \underline{60.60\ppm0.53}   & \underline{82.40\ppm0.17}     &  \textbf{83.97\ppm0.18} &   \underline{74.59\ppm0.21}  \\
ATFS-ViT (ours) & T2T-ViT-14 & 21.5M &  \textbf{71.99\ppm0.10}  &  \textbf{60.67\ppm0.10} &   82.35\ppm0.70	 &  \underline{83.59\ppm0.65}  	 &  \textbf{74.65\ppm0.24}   \\
\bottomrule
\end{tabular}
}
\end{center}
\end{table*}

\begin{table*}[t]
\begin{center}
\caption{ Our proposed method performance on different domains of the TerraIncognita~\cite{beery2018recognition} dataset. Mean and Standard Deviation are reported across three runs. The best and second best average is in \textbf{bold} and \underline{underlined} fonts, respectively.}
\label{tab:ablation_terra}
\adjustbox{max width=\textwidth}{
\begin{tabular}{lccccccc}
\toprule \noalign{\smallskip}
\textbf{Method} & \textbf{Backbone} & \textbf{\#\,of Params} &  \textbf{location\_100} & \textbf{location\_38} & \textbf{location\_43} & \textbf{location\_46} & \textbf{Average} \\
\noalign{\smallskip}
\midrule
\noalign{\smallskip}
ERM-ViT  & DeiT-Small & 22M & 50.2\ppm1.4 & 30.6\ppm0.9& 53.2\ppm0.2 & 39.6\ppm1.0 &43.4\ppm0.5  \\
SDViT~\cite{sultana2022self} & DeiT-Small & 22M & 55.9\ppm1.7   &       31.7\ppm2.6    &      52.2\ppm0.3  &        37.4\ppm0.6  &   44.3\ppm1.0   \\
TFS-ViT (ours) & DeiT-Small & 22M &  58.70\ppm1.43   &   40.85\ppm1.92    &  53.72\ppm0.47   &  41.14\ppm0.22 & 48.60\ppm0.61     \\
ATFS-ViT (ours) & DeiT-Small & 22M &  57.47\ppm2.61   &  36.45\ppm0.47     &   52.19\ppm0.56 	 &   38.11\ppm0.62 	 & 46.06\ppm0.70    \\
\midrule
ERM-ViT  & T2T-ViT-14 &  21.5M &52.5\ppm1.7      &    43.0\ppm1.3   &       53.7\ppm1.1  &        \underline{43.0\ppm1.6}         &  48.1\ppm0.2    \\
SDViT\cite{sultana2022self} & T2T-ViT-14 & 21.5M &  57.2\ppm2.9       &   \underline{45.4\ppm2.4}     &     \underline{57.7\ppm0.8}  &        41.9\ppm0.4    & 50.6\ppm0.8   \\
TFS-ViT (ours) & T2T-ViT-14 & 21.5M &  \underline{59.62\ppm1.44}    &    \textbf{47.40\ppm1.28}   & \textbf{58.19\ppm0.34}     &  41.84\ppm0.89 &   \textbf{51.76\ppm0.54} \\
ATFS-ViT (ours) & T2T-ViT-14 & 21.5M &  \textbf{60.32\ppm0.51}  &  43.72\ppm1.34 &    56.39\ppm0.23	 &  \textbf{44.38\ppm0.93} 	 &  \underline{51.20\ppm0.43}   \\
\bottomrule
\end{tabular}
}
\end{center}
\end{table*}

\begin{table*}[t]
\begin{center}
\caption{Our proposed method performance on different domains of the DomainNet~\cite{peng2019moment} dataset. Mean and Standard Deviation are reported across three runs. The best and second best average is in \textbf{bold} and \underline{underlined} fonts, respectively.}
\label{tab:ablation_dn}
\adjustbox{max width=\textwidth}{
\begin{tabular}{lcccccccccc}
\toprule \noalign{\smallskip}
\textbf{Method} & \textbf{Backbone} & \textbf{\#\,of Params} &  \textbf{Clipart} & \textbf{Infograph} & \textbf{Painting} & \textbf{Quickdraw} & \textbf{Real}  & \textbf{Sketch} & \textbf{Average} \\
\noalign{\smallskip}
\midrule
\noalign{\smallskip}
ERM-ViT  & DeiT-Small & 22M &62.9\ppm0.2   &       23.3\ppm0.1    &      53.1\ppm0.2   &       15.7\ppm0.1    &      65.7\ppm0.1    &      52.4\ppm0.2 & 45.5\ppm0.0  \\
SDViT~\cite{sultana2022self} & DeiT-Small & 22M & 63.4\ppm0.1   &       22.9\ppm0.0     &     53.7\ppm0.1    &      15.0\ppm0.4      &    67.4\ppm0.1       &   52.6\ppm0.2  &  45.8\ppm0.0 \\
TFS-ViT (ours) & DeiT-Small & 22M & 64.85\ppm0.14    &       23.51\ppm0.11     &     53.60\ppm0.15    &      16.60\ppm0.00      &    67.61\ppm0.20       &   53.44\ppm0.23  &  46.60\ppm0.06 \\
ATFS-ViT (ours) & DeiT-Small & 22M & 64.36\ppm0.30   &       23.37\ppm0.19     &     53.35\ppm0.21    &       15.80\ppm0.00        &     67.39\ppm0.04       &    52.79\ppm0.16   &  46.18\ppm0.07 \\
\midrule
ERM-ViT  & T2T-ViT-14 &  21.5M & \underline{67.0\ppm0.3}       &   \underline{25.2\ppm0.2}   &       55.3\ppm0.3  &        15.3\ppm0.2   &       \textbf{70.3\ppm0.1}     &     \underline{55.9\ppm0.2}  & 48.1\ppm0.1  \\
SDViT\cite{sultana2022self} & T2T-ViT-14 & 21.5M & \textbf{67.6\ppm0.2} &     25.0\ppm0.2 &    \underline{55.8\ppm0.4}   &       15.2\ppm0.3 &     \underline{70.0\ppm0.1}   &       \textbf{55.9\ppm0.1}  & \underline{48.2\ppm0.2} \\
TFS-ViT (ours) & T2T-ViT-14 & 21.5M & 66.15\ppm0.31 &    \textbf{25.42\ppm0.04} &    \textbf{56.11\ppm0.49}   &       \textbf{17.04\ppm0.21} &     69.54\ppm0.18   &        55.75\ppm0.42  & \textbf{48.34\ppm0.13} \\
ATFS-ViT (ours) & T2T-ViT-14 & 21.5M & 66.31\ppm0.30 &     24.74\ppm0.51 &    55.44\ppm0.61   &       \underline{16.82\ppm0.81} &     69.08\ppm0.40   &       55.27\ppm0.24  & 47.94\ppm0.21 \\
\bottomrule
\end{tabular}
}
\end{center}
\end{table*}

